\documentclass{article}

\usepackage{arxiv}

\usepackage[utf8]{inputenc} % allow utf-8 input
\usepackage[T1]{fontenc}    % use 8-bit T1 fonts
\usepackage{url}            % simple URL typesetting
\usepackage{booktabs}       % professional-quality tables
\usepackage{amsfonts}       % blackboard math symbols
\usepackage{nicefrac}       % compact symbols for 1/2, etc.
\usepackage{microtype}      % microtypography
\usepackage{lipsum}		% Can be removed after putting your text content
\usepackage{graphicx}
\usepackage[numbers]{natbib}
\usepackage{doi}

\usepackage{amsmath}
\usepackage{amsthm}
\PassOptionsToPackage{table}{xcolor}
\usepackage{tikz}
\usepackage{fontawesome} % For thumbs up or down
\usepackage{pifont} % For checkmark
\usepackage{wasysym} % For male/female
\usepackage[capitalize, noabbrev, nameinlink]{cleveref}
\usepackage{colortbl} % Column colors
\usepackage{pgfplots}
\pgfplotsset{compat=1.10}
\usepgfplotslibrary{fillbetween}
\usetikzlibrary{matrix}
\usetikzlibrary{patterns}
\usetikzlibrary{backgrounds}
\usepackage{subcaption}
\usepackage{multirow}
\usepackage{tcolorbox}

\pgfplotsset{scaled y ticks=false}

\newcommand{\personSymbol}[1]{%
    \tikz[scale=0.1]{
        \begin{scope}
            \clip (-3, -0.05) rectangle (1.75, 1.5);
            \draw[fill=#1] (0,0) circle(1.5);
            \draw[fill=#1] (-1.5,0) -- (1.5,0);
        \end{scope}
        \draw[fill=#1] (0, 1.75) circle (0.825);
        \draw[opacity=0] (2.0, 1.75) circle (0.825);
    }%
}
\newcommand{\thumbsUp}{%
    {\color{green!60}\faThumbsUp}%
}
\newcommand{\thumbsDown}{%
    {\color{red!60}\faThumbsDown}%
}
\newcommand{\cmark}{\textcolor{black}{\ding{51}}}%
\newcommand{\xmark}{\textcolor{black!15}{\ding{55}}}%

\newcommand{\PreserveBackslash}[1]{\let\temp=\\#1\let\\=\temp}
\newcolumntype{C}[1]{>{\PreserveBackslash\centering}p{#1}}
\newcolumntype{R}[1]{>{\PreserveBackslash\raggedleft}p{#1}}
\newcolumntype{L}[1]{>{\PreserveBackslash\raggedright}p{#1}}

\newtheorem{assumption}{Assumption}

\newcommand{\ind}{\perp\!\!\!\!\perp} 

\newcommand{\bftab}{\fontseries{b}\selectfont}

\newcommand{\plus}{\textcolor{white!0}{$+$}}%
\newcommand{\minus}{\textcolor{black}{$-$}}%

\title{Metalearners for Ranking Treatment Effects}

\author{Toon ~Vanderschueren\thanks{Correspondence to: \href{mailto:toon.vanderschueren@gmail.com}{\texttt{toon.vanderschueren@gmail.com}}. Work conducted while the first author was an intern at Booking.com.} \\
	KU Leuven, University of Antwerp \\ 
	\And
	Wouter Verbeke \\
	KU Leuven\\
	\And 
        Felipe Moraes \\
        Booking.com
        \And 
        Hugo Manuel Proença \\
        Booking.com \\
}

\renewcommand{\shorttitle}{Metalearners for Ranking Treatment Effects}

\hypersetup{
    pdftitle={Metalearners for Ranking Treatment Effects},
    pdfsubject={q-bio.NC, q-bio.QM}, % TODO
    pdfauthor={Toon ~Vanderschueren, Wouter ~Verbeke, Felipe ~Moraes, Hugo ~Proença},
    pdfkeywords={Causal Inference, Treatment Effect Estimation, Learning to Rank},
}

\begin{document}
\maketitle

\begin{abstract}
    Efficiently allocating treatments with a budget constraint constitutes an important challenge across various domains. In marketing, for example, the use of promotions to target potential customers and boost conversions is limited by the available budget. While much research focuses on estimating causal effects, there is relatively limited work on learning to allocate treatments while considering the operational context. Existing methods for uplift modeling or causal inference primarily estimate treatment effects, without considering how this relates to a profit maximizing allocation policy that respects budget constraints. The potential downside of using these methods is that the resulting predictive model is not aligned with the operational context. Therefore, prediction errors are propagated to the optimization of the budget allocation problem, subsequently leading to a suboptimal allocation policy. We propose an alternative approach based on learning to rank. Our proposed methodology directly learns an allocation policy by prioritizing instances in terms of their incremental profit. We propose an efficient sampling procedure for the optimization of the ranking model to scale our methodology to large-scale data sets. Theoretically, we show how learning to rank can maximize the area under a policy's incremental profit curve. Empirically, we validate our methodology and show its effectiveness in practice through a series of experiments on both synthetic and real-world data.
\end{abstract}

% keywords can be removed

\section{Introduction}

Decision-makers need to deal with uncertainty regarding the consequences of their decisions. An increasingly popular paradigm to address this challenge is the prediction-optimization framework. In a first \textit{prediction} stage, data is used to estimate the effect of possible actions. In a second \textit{optimization} stage, these predictions are integrated in an optimization problem with the aim of assigning personalized treatment recommendations, i.e., allocating treatments to instances to optimize an objective function, while satisfying operational constraints. These problems are common in various domains: e.g., marketing \citep{albert2022commerce}, healthcare \citep{savachkin2012dynamic}, maintenance \citep{vanderschueren2023optimizing}, or policy design \citep{bhattacharya2012inferring} (see \cref{tab:intro_examples} for some examples). We focus on a specific class of treatment recommendation problems where instances need to be prioritized for treatment (e.g., recommending whom to treat). Our goal is to learn a treatment policy that prioritizes the optimal instances for treatment. A key part of this problem is estimating each instance's response to a treatment--i.e., its treatment effects--from data using causal inference.

\begin{table*}[t]
    \centering
    \setlength{\tabcolsep}{4pt}
    \begin{tabular}{L{6pt}L{60pt}|L{95pt}L{85pt}L{85pt}L{85pt}}
    \toprule
        & \textbf{\textit{Application}}
        & \textbf{Allocation problem} & \textbf{Treatment outcome} & \textbf{Objective} & \textbf{Constraints} \\
    \midrule
        \faPercent & \textit{Marketing}    & Targeted advertising     & Conversion & Incremental sales          & Marketing budget \\ 
        \faMedkit & \textit{Healthcare}  & Pandemic response        & Infection & Mortality reduction       & Vaccine supply \\
        \faWrench & \textit{Maintenance} & Preventive maintenance   & Failure rate & Asset uptime & Available technicians \\
        \faMoney & \textit{Policy design}    & Targeted subsidies     & Bed net purchase & Malaria prevention          & Public/policy budget \\
    \bottomrule
    \end{tabular}
    \vspace{2pt}
    \caption{\textit{Treatment Allocation Examples.} We highlight several applications where treatment allocation is required, characterized by (1) an estimated treatment effect, (2) an optimization objective, and (3) operational constraints. In \textit{marketing}, the goal is to target customer segments and drive conversions, while adhering to budget constraints. In \textit{healthcare and epidemiology}, optimal vaccine allocation during pandemics aims to minimize population mortality, subject to vaccine supplies. In \textit{maintenance}, technicians are assigned to maintain assets, prevent failures, and maximizing operational uptime. Finally, in \textit{economics and policy design}, subsidy usage needs to be optimized by efficiently allocating public funds and maximizing health care impact.}
    \label{tab:intro_examples}
    % \vspace{-5pt}
    \noindent\rule{\textwidth}{0.1pt}
    % \vspace{-10pt}
\end{table*}

% Effect estimation
% Predict-then-optimize
\paragraph{Prediction-Focused Learning: Effect Estimation}
A common approach to tackle treatment allocation problems is to first \textit{predict} the effect of an action for each instance. To this aim, causal inference can support many decision-making problems: by analyzing the causal effect of past decisions, future decisions can be optimized. A common approach is to first \textit{estimate the causal effect} of possible decisions using methods for treatment effect estimation. For example, in marketing, to estimate how different customers would respond to a marketing incentive. The effect estimates can be integrated in an \textit{optimization} problem to make the final decisions regarding treatment allocation (e.g., to target a specific customer segment). This approach has been adopted to aid decision-making by a variety of technology and e-commerce companies \citep{diemert2018large, goldenberg2020free, syrgkanis2021causal}.

% Effect ranking
% Predict-and-optimize, decision-focused learning
\paragraph{Decision-Focused Learning: Effect Ranking}
Recent work, referred to as \textit{decision-focused learning}, aims to integrate the learning and optimization steps. This approach recognizes that the predictive task (i.e., estimating treatment effects) is only part of a larger optimization problem (i.e., allocating treatments). By integrating the predictive model in the larger optimization pipeline, decision-focused learning aims to learn a predictive model that results in better performance for the downstream task \cite{mandi2023decision}. The key idea is to align the construction of the predictive model with the optimization task. 

We analyze a common type of treatment allocation problem, where treatments are allocated to the instances with the largest treatment effect. In these settings, we argue that directly learning an \textit{effect ranking} might be more useful than \textit{effect estimation}. As operational constraints might prevent decision-makers from treating every instance, we require knowing how to prioritize instances based on their treatment effect. Compared to independently estimating each instance's effect, we argue that directly learning the ranking across instances may yield better results. Because \textit{effect estimation} prioritizes accurate and well-calibrated effect estimates, it overlooks the estimates' ranking and resulting decision quality (i.e., estimation and optimization are not aligned). While successful when predictions are perfect, this misalignment can result in suboptimal decision-making in reality. Conversely, our work demonstrates that \textit{effect ranking} can directly optimize the quality of the treatment assignment (i.e., the ranking objective is perfectly aligned with the decision-making task). We contrast both approaches in \cref{tab:toy_example}. Additionally, empirical risk minimization only guarantees model generalization for the specific objective that was optimized for \cite{betlei2021uplift}, further motivating us to find objectives that are aligned to the final optimization problem.

\paragraph{Contributions}

This work proposes decision-focused learning framework for treatment recommendation problems. We formalize the class of problems that can be tackled using effect ranking and discuss the underlying assumptions. We describe how learning to rank can be used for this task and propose different \textit{causal metalearners for ranking effects}. 
Our contributions are as follows. \text{(1)} Conceptually, we formalize treatment allocation problems that can be solved using ranking and analyze the underlying assumptions (see \cref{sec:problem_formulation}). \text{(2)} Methodologically, we propose different metalearners for ranking treatment effects, based on pairwise and listwise ranking objectives that scale efficiently to large-scale data sets. We show how our proposed listwise objective directly optimizes the policy's area under the Qini curve (see \cref{sec:methodology}). \text{(3)} Empirically, we compare our proposed effect ranking with effect estimation using synthetic and real-world data sets (see \cref{sec:empirical_results}).

\begin{table*}[!t]
    \centering
    {\renewcommand{\arraystretch}{1}}  % Vertical spacing 
    \setlength{\tabcolsep}{7pt}  % Horizontal spacing
    \begin{tabular}
    {R{110pt}C{30pt}C{30pt}C{30pt}C{30pt}C{30pt}C{30pt}cc}
    \toprule
         & \multicolumn{6}{c}{\textbf{Instances (e.g., customers)}} & \multicolumn{2}{c}{\textbf{Performance}} 
    \\
     & \personSymbol{green!50} & \personSymbol{red!50} & \personSymbol{blue!50} & \personSymbol{purple!50} & \personSymbol{yellow!50} & \personSymbol{orange!50} & MSE & Ranking \\
    \cmidrule(lr){2-7} \cmidrule(r){8-9}
        \rowcolor{gray!10} \textbf{\textit{True treatment effect}} & \textit{0.20} & \textit{0.17} & \textit{0.16} & \textit{0.15} & \textit{0.11} & \textit{0.10} & --- & --- \\
    \midrule
        \textbf{Effect estimation: model 1} & 0.22 & 0.15 & 0.16 & 0.17 & 0.10 & 0.11 & \thumbsUp & \thumbsDown \\
        \textbf{Effect ranking: model 2} & 0.25 & 0.22 & 0.20 & 0.15 & 0.10 & 0.05 & \thumbsDown & \thumbsUp \\
    \bottomrule
    \end{tabular}
    \vspace{2pt}
    \caption{\textit{Comparing Pointwise Estimation and Effect Ranking for Treatment Allocation.} This illustrative example shows the true treatment effect for several instances, ordered from largest to smallest. The first model estimates each instance's treatment effect fairly accurately in terms of MSE. However, the ranking of instances based on these estimates differs significantly from their true order, which will result in suboptimal treatment allocation when only some instances can be treated. For the second model, we observe the opposite scenario: estimates are poor in terms of MSE, but their ranking respects the true order.}
    \label{tab:toy_example}
    % \vspace{-5pt}
    \noindent\rule{\textwidth}{0.1pt}
    % \vspace{-10pt}
\end{table*}

\section{Related Work}

In the following, we discuss two areas of related work. First, causal inference and, more specifically, estimating treatment effects. Second, learning policies for treatment recommendation.

\subsection{Prediction-Focused Learning: Effect Estimation}

Understanding the impact of an action on an instance or individual is crucial in a variety of domains where personalized decision-making is valuable, such as marketing, healthcare, or education. Central to this is causal inference: using data to draw conclusions regarding causal relationships. Especially relevant to our work is treatment (or causal) effect estimation, where the aim is to learn how some treatment, action, or intervention will affect an instance's outcome of interest. For a comprehensive review, we refer to Zhang et al. \cite{zhang2021unified}. In the context of marketing, using machine learning (ML) for treatment effect estimation is generally referred to as uplift modeling, as discussed in several surveys \citep{devriendt2018literature, olaya2020survey, zhang2021unified}. 

Specialized methods have been proposed for estimating treatment effects. First, general strategies exist for learning these effects. Causal metalearners are modeling frameworks for effect estimation, compatible with various ML algorithms \citep{betlei2018upliftddr, kunzel2019metalearners, curth2021nonparametric}. Relatedly, response transformation approaches transform an instance's outcome so that it can be modeled using a standard classifier \citep{kane2014mining, devriendt2018literature}. Second, ML algorithms have been adapted for effect estimation, such as decision trees \citep{rzepakowski2012decision} or random forests \citep{soltys2015ensemble, guelman2015uplift}.

% Optimization perspective 
% 1. Profit-driven/cost-sensitive learning
While uplift modeling has traditionally focused on optimizing conversion, practitioners often seek to optimize other metrics related to their business and operational context. Recent work explores cost-sensitive or profit-driven uplift modeling, where the aim is to estimate and maximize profit and cost resulting from targeting policies \citep{verbeke2020foundations, devriendt2021you, gubela2021uplift, verbeke2023or}. For example, the Incremental Profit per Conversion (IPC) has been proposed as a response transformation approach for incremental profit \citep{proencca2023incremental}. 

All work in this category aims to \textit{estimate the effect} of a treatment (e.g., a customer's incremental conversion probability as a result of receiving a marketing incentive). As discussed in the introduction, these estimates can be used to design a treatment allocation policy (e.g., by targeting customers with a large estimatd effect). In practice, operational constraints (such as budget limitations) may call for more complex optimization procedures \citep[][]{zhao2019unified, tu2021personalized, ai2022lbcf, albert2022commerce}. The resulting treatment allocation problem requires solving a constrained optimization problem using the effect estimates as input. 

% 2. Ranking
\subsection{Decision-Focused Learning: Effect Ranking}
% More general 
% Predict-and-optimize / decision-focused learning

As argued before, the prediction-focused approach may suffer from a misalignment between prediction and optimization, leading to suboptimal treatment decisions. Additionally, empirical risk minimization only guarantees model generalization for the specific objective that was optimized \cite{betlei2021uplift}. Decision-focused learning aims to align the two phases by using an integrated approach and directly optimizing a predictive model for the final optimization problem. Related to this insight, a handful of methods have recently been proposed for treatment recommendation that aim to learn a ranking of instances in terms of their treatment effect. Although these methods were originally proposed in the context of uplift modeling--framing targeted marketing as a ranking problem--they are more generally applicable. Finally, it has been noted that any score--even non-causal estimands--can be used to prioritize instances for treatment, as long as it is a good proxy for the treatment effect's magnitude \citep{fernandez2022learning, athey2023machine}.

\cref{tab:literature_table} highlights related methods for ranking treatment effects, describing the metalearners and optimization strategies that were used, where we follow the literature on learning to rank \citep{cao2007learning, liu2009learning}. The first approach, \textit{pointwise} ranking, relies on an estimate of the treatment effect and corresponds to prediction-focused learning. The second approach, \textit{pairwise} ranking, aims to predict the relative ranking between instances over all pairs of instances. The final approach, \textit{listwise} ranking, optimizes the ranking across all instances in the ranking simultaneously. In the literature on learning to rank, pairwise and listwise ranking approaches have surpassed pointwise approaches, with listwise methods typically performing best \citep{liu2009learning}.

Most existing approaches for ranking effects rely on alternative objective functions that aim to integrate prediction and optimization using Lagrangian duality \citep{du2019improve, zou2020heterogeneous, zhou2023direct} or gradient estimation techniques \citep{yan2023end}. Alternatively, the causal profit ranker \citep{verbeke2023or} ranks instances in a post-processing stage using pointwise estimates of the expected conversion More advanced pairwise \citep{betlei2021uplift} and listwise \citep{devriendt2020learning} learning to rank have also been explored in this context. Conversely, our work explores pointwise, pairwise, and listwise objectives, as well as a wide variety of metalearners.

\begin{table}[t]
    \centering
    \begin{tabular}{R{24pt}|C{24pt}C{24pt}C{24pt}|C{24pt}C{24pt}C{24pt}C{24pt}C{24pt}C{24pt}} % 
    \toprule
                        & \multicolumn{3}{c|}{\textbf{Objective}}                              & \multicolumn{6}{c}{\textbf{Metalearners}} \\
        {\textit{Ref.}} & \textit{Point} & \textit{Pair} & \textit{List} & \textit{Z} & \textit{S} & \textit{T} & \textit{X} & \textit{DR} & \textit{R} \\
    \midrule
        \cite{du2019improve}            & \cmark & \xmark & \xmark & \xmark & \cmark & \xmark & \xmark & \xmark & \xmark \\
        \cite{zou2020heterogeneous}     & \cmark & \xmark & \xmark & \xmark & \cmark & \xmark & \xmark & \xmark & \cmark \\
        \cite{devriendt2020learning}    & \cmark & \xmark & \cmark & \cmark & \xmark & \xmark & \xmark & \xmark & \xmark \\
        \cite{betlei2021uplift}         & \cmark & \cmark & \xmark & \xmark & \cmark & \xmark & \xmark & \xmark & \xmark \\
        \cite{verbeke2023or}            & \cmark & \xmark & \xmark & \xmark & \cmark & \cmark & \xmark & \xmark & \xmark \\
        \cite{zhou2023direct}           & \cmark & \xmark & \xmark & \xmark & \cmark & \xmark & \xmark & \xmark & \xmark \\
    \midrule
        % \rowcolor{gray!10} 
        {Ours}                        & \cmark & \cmark & \cmark & \cmark & \cmark & \cmark & \cmark & \cmark & \cmark \\
    \bottomrule
    \end{tabular}
    \vspace{5pt}
    \caption{\textit{Literature Table.} We categorize related work on effect ranking by differentiating between different (1) ranking approaches (point-, pair-, and listwise) and (2) metalearners.}
    \label{tab:literature_table}
    % \vspace{-10pt}
    \noindent\rule{\linewidth}{0.1pt}
    \vspace{-10pt}
\end{table}

General methodologies have been proposed for learning a treatment policy, directly mapping an instance's characteristics to a recommended treatment \cite{dudik2011doubly, zhang2012estimating, laber2015tree, luedtke2016super, kitagawa2018should, athey2021policy}. In the context of our work, these can be seen as pointwise approaches, as they do not consider the ranking structure of the optimization task.

\section{Problem Formulation}
\label{sec:problem_formulation}

In this work, we aim to learn a treatment policy that prioritizes instances for treatment to maximize the aggregate effect. As opposed to the existing work, which assumes a problem formulation implicitly, we explicitly formalize our problem setting. In doing so, we reveal the underlying assumptions required for our approach.

\subsection{Notation and Optimization Problem}

Let an instance (e.g., a customer) be described by a tuple $(x_i, t_i, y_i)$, representing covariates $X \subset \mathbb{R}^d$, an administered treatment $t \in \{0, 1\}$, and the outcome to be optimized $Y \subset \mathbb{R}$. We denote the potential outcome $Y$ associated with a treatment $t$ as $Y(t)$ and an instances $i$'s treatment effect as $\tau_i = Y_i(1) - Y_i(0)$ (e.g., a customer's incremental conversion probability resulting from receiving a discount). We aim to learn a policy $\pi$ that assigns treatments to instances and maximizes the overall treatment effect, while respecting possible operational constraints. At test time, we assume $n$ instances can be treated, subject to a treatment budget $B$ with $B \leq n$. This yields the following optimization problem:
\begin{equation*}
\begin{aligned}
\max_{\mathbf{t_i}} \quad & \sum_{i=1}^{n} \tau_i(t_i) \\
\textrm{s.t.} \quad & \sum_{i=1}^n t_i \leq B \\
  &\; t_i \in \{0, 1\} \;\; \forall \; i \in \{0, \dots, n\}    \\
\end{aligned}
\end{equation*}
At test time, treatment effects $\tau$ are unknown and need to be estimated. To this end, we assume access to a historical data set $\mathcal{D} = \{x_i, t_i, y_i(t_i)\}_{i=1}^m$ describing past treatment decisions and the resulting outcomes. These data can be used to estimate the conditional average treatment effect (CATE): $\hat{\tau} = \mathbb{E}(Y(1) - Y(0) \mid X = x)$.

\subsection{Assumptions} We make several assumptions regarding the causal structure of the data and the operational constraint. 
To estimate the causal effect from historical data, we require the standard assumptions for identifiability in causal inference \citep{rubin1974estimating} (see \cref{sec:assumptions_app} for a more extensive discussion). Additionally, we more formally define the operational constraint. Although we assume that not all instances can be treated and, thus, instance prioritization is required, we assume that the exact budget is not known to the decision-maker a priori. More formally, we state there is no information regarding the budget $B$ a priori:
\begin{assumption}[Operational constraint]
    We assume the exact budget is unknown, but the expectation is uniformly distributed among $\{1, \dots, n\}$: $B \sim \mathcal{U}(1, n)$.
\end{assumption}
We discuss how alternative assumptions regarding the budget would affect our proposed solution below.

\subsection{Evaluating a Treatment Policy}
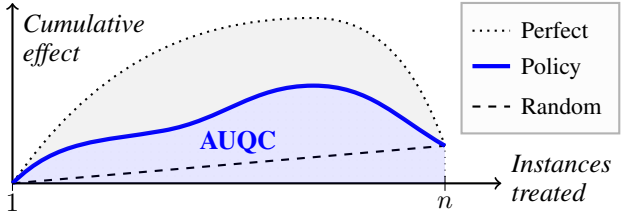
\begin{figure}[t]
    \centering
    \tikzset{
      smallpath/.pic={
        \draw (0, 0.1) to (0.5, 0.1);
      }
    }
    
    \begin{tikzpicture}[scale=1]
    % Axes:
    \draw[black, thick, ->] (0, 0) to (6.5, 0);
    \draw[black, thick, ->] (0, 0) to (0, 2.4);
    \node[draw, align=center, draw=none] at (0, -0.25) {$1$};
    \draw[black] (0, -0.1) to (0, 0.1);
    \node[draw, align=center, draw=none] at (5.75, -0.25) {$n$};
    \draw[black, thin, dotted] (5.75, 0) to (5.75, 0.5);
    \draw[black] (5.75, -0.1) to (5.75, 0.1);
    \node[draw, align=left, draw=none] at (0.95, 1.9) {\textit{Cumulative}\\ \textit{effect}};
    \node[draw, align=left, draw=none] at (7.3, 0.075) {\textit{Instances}\\ \textit{treated}};

    % Random curve:
    \draw[black, thick, dashed, name path=random] (0, 0) to (5.75, 0.5);

    % Perfect curve:
    \draw [black, thick, dotted, name path=perfect] (0, 0) to[out=55, in=180] (4, 2.2) to[out=0, in=110] (5.75, 0.5);

    % Model curve:
    \draw [blue, ultra thick, name path=model] (0, 0) to[out=45, in=190] (2, 0.7)  to[out=180+190, in=180] (4, 1.3) to[out=0, in=150] (5.75, 0.5);

    % Invisible curve at zero for fillbetween:
    \draw [name path=xaxis] (0, 0) to[] (5.75, 0);
    
    \tikzfillbetween[of=xaxis and model]{blue, opacity=0.1};
    \tikzfillbetween[of=xaxis and model]{pattern=dots, opacity=0.1};
    \tikzfillbetween[of=perfect and model]{gray, opacity=0.1};

    \node[blue, anchor=center] at (3, 0.55) {\textbf{AUQC}};

    % Legend
    \matrix[matrix of nodes, draw=gray!60, below left, fill=gray!2, row sep=0pt, column sep=0pt, thick,
    every node/.style={anchor=base west}, font=\small,
    nodes={rectangle, draw=none, minimum height=14pt},
    column 1/.style={anchor=base center}
    column 2/.style={anchor=base west}] 
    at (current bounding box.north east) {
        \pic[black, thick, dotted]{smallpath}; & Perfect\\
        \pic[blue, ultra thick]{smallpath}; & Policy\\
        \pic[black, thick, dashed]{smallpath}; & Random\\
    };
    \end{tikzpicture}
    \vspace{-5pt}
    \caption{\textit{Evaluating a Treatment Allocation Policy.} We compare targeting policies using a Qini curve, depicting the cumulative total effect of a policy for a number of treated instances, summarized by the area under the Qini curve (AUQC).}
\vspace{-5pt}
\noindent\rule{\linewidth}{0.1pt}
\vspace{-20pt}
\label{fig:qini_illustration}
\end{figure}

We assess the quality of a proposed policy using the Qini curve, illustrated in \cref{fig:qini_illustration}. This curve shows the cumulative total effect of a policy for a number of treated instances \cite{gutierrez2017causal, sverdrup2023qini}. Given that we assume no information regarding the budget, we measure a policy's overall quality using the area under the Qini curve (AUQC), quantifying the total cumulative effect over the entire ranking. Formally, we define the (hypothetical) AUQC as
\begin{equation}
    \text{AUQC} = \sum_{k=1}^n \sum_{i=1}^k \tau_{i}
\label{eq:auqc}
\end{equation}
with $\tau_i$ the effect of the instance at position $i$ in the ranking. The normalized AUQC is obtained by comparing it with the expected AUQC of a random ranking and AUQC of a perfect ranking. 
Typically, the normalized AUQC ranges between zero and one $\in [0, 1]$, though a worse than random policy with AUQC $<0$ is also possible.
Because effects $\tau$ are not observed in reality, Qini curves need to be estimated from data on past treatment allocations \cite{devriendt2020learning, sverdrup2023qini}.

\section{Methodology}
\label{sec:methodology}

Given the problem setup described above, we now present our proposed methodology, which essentially learns a ranking (or sorting) of each instance's treatment effect. The optimality of this solution can be seen as follows. If only one instance can be treated (i.e., $B=1$), the optimal solution is to assign the treatment to the instance with the largest treatment effect $\tau_i$. Given an unknown budget and uniform expectation regarding this budget, the optimal solution is then to rank all instances by their treatment effect $\tau_i$ and assign treatments to the top instances until the budget runs out. Therefore, our goal is to predict an optimal ordering or assignment policy $\pi \in \Pi_n$ that permutes the test instances $\{1, \dots, n\}$ to the optimal ordering based on descending treatment effects $\tau$.

% Envisioned solution
As previously discussed, most existing approaches first estimates the effects $\hat{\tau}$ and then rank these estimates. However, as discussed above and in \cref{tab:toy_example}, this approach has two drawbacks. First, the estimator's objective is not aligned with the optimization task, possibly resulting in suboptimal decisions \citep{mandi2023decision}. Second, the resulting model is only guaranteed to generalize for the predictive objective that was used \citep{betlei2021uplift}. These issues motivate us to directly learn a ranking policy $\pi$ based on instance characteristics $X$,  which requires addressing two challenges. First, to find an objective that optimizes a ranking of instances instead of a pointwise estimate (see \cref{ssec:objectives}). Second, the ranking needs to be based on the treatment effect $\tau$, which is not observed. Therefore, we extend metalearners for effect estimation to ranking (see \cref{ssec:metalearners}).

\subsection{Optimizing a Ranking Objective}
\label{ssec:objectives}

In this section, we explore approaches for optimizing a ranking. We discuss pointwise, pairwise, and listwise approaches. We propose a listwise objective that optimizes the policy's AUQC directly. Additionally, we propose a sampling strategy to improve the efficiency of our proposed ranking objectives.

\subsubsection{Ranking objectives}
We describe three objectives for learning a ranking policy $\pi$ that can be used by ML algorithms. 

\paragraph{Pointwise} The first approach, used by most existing work, is to learn a pointwise estimate of the effect. As the treatment effect itself is never observed, this corresponds to either learning the observed outcome $y(t)$ or a transformed outcome, depending on the metalearner used (see \cref{ssec:metalearners}). In this work, we use the mean squared error as a pointwise objective to learn the estimand: 
\begin{equation}
    \mathcal{L_\text{Point}}(y, \hat{y}) = \frac{1}{n} \sum_{i=1}^n \left(y_i - \hat{y}_i\right)^2
    .
\end{equation}
While other objectives are possible \citep[e.g.][]{zhou2023direct}, pointwise approaches by definition ignore the instance ranking resulting from the point estimates. This motivates our exploration of alternative objectives.

\paragraph{Pairwise}
The first ranking approach is the pairwise approach. The idea is to predict, for each pair of instances, how both instances are ranked respective to each other. If all pairs are ranked correctly, the overall ranking will also be correct.
We build upon the approach proposed for RankNet \cite{burges2005learning}.
Before we define the pairwise objective, we define the pairwise outcome $y_{i,j}$ that specifies whether instance $i$ or $j$ should be ranked higher, for each pair of instances $i$ and $j$: 
\begin{equation*}
    y_{i,j} =
    \begin{cases}
      1   & \text{if $y_i \geq y_j$} \\
      0   & \text{if $y_i < y_j$}.
    \end{cases}
\end{equation*}
Similarly, we define a smooth pairwise prediction $\hat{y}_{i,j}$, combining two instances' predictions $\hat{y}_i$ and $\hat{y}_j$ as follows:
\begin{equation}
    \hat{y}_{i,j} = \frac{1}{1 + \exp\big(\!-\!\sigma (\hat{y}_i - \hat{y}_j)\big)}
    ,
\label{eq:pairwise_score}
\end{equation}
where the sigmoid parameter $\sigma$ controls the smoothness of the comparison. In the extreme $\sigma = \infty$, this becomes a step function. 
This way, we define pairwise ranking as a binary classification task with the pairwise cross-entropy loss defined as: 
\begin{equation}
    \mathcal{L_\text{Pair}}(y, \hat{y}) = \sum_{i=1}^n \sum_{j=1}^n - y_{i,j} \log \left( \hat{y}_{i,j} \right) - (1 - y_{i,j}) \big(1 - \log \left( \hat{y}_{i,j} \right)\! \big)
    .
\end{equation}
Other pairwise objectives exist \cite[e.g.][]{betlei2021uplift} which can be applied to our approach, though we consider this outside the scope of this work. 

\paragraph{Listwise}
A drawback of the pairwise approach is that it overlooks the relative importance of correctly classifying one pair on the listwise ranking quality. To address this issue, the LambdaMART objective \cite{burges2010lambdamart} adds a weight $\text{NDCG}_{i,j}$ to the pairwise objective, reflecting the increase in normalized discounted cumulative gain (NDCG, see below) achieved by swapping that pair:
\begin{align}
    \mathcal{L_\text{List}}(y, \hat{y}) = \sum_{i=1}^n \sum_{j=1}^n \bigg(  - y_{i,j} \log \left( \hat{y}_{i,j} \right)
    - (1 - y_{i,j}) \big(1 - \log \left( \hat{y}_{i,j} \right)\! \big) \bigg) 
    \textcolor{orange}{\Delta \text{NDCG}_{i,j}}.
    \notag 
\end{align}

\subsubsection{The AUQC as a specific instance of the NDCG}

The normalized discounted cumulative gain (NDCG) is a class of metrics measuring the quality of a ranking \citep{wang2013theoretical}. Formally, we define a ranking $\pi$, with $\pi_i$ representing the $i$'th instance in the ranking. An instance's gain $g_i$ represents its value independent of its position in the ranking (e.g., the treatment effect $\tau$). The NDCG decreases the gain for lower ranks, reflecting their decreasing importance, by applying a discount function $d(i)$ to the instance's gain $g_i$. The discounted cumulative gain (DCG) is the sum of all discounted gains over the ranking $\text{DCG} = \sum_{i=1}^n d(i) g_{\pi_i}$. Finally, the normalized discounted cumulative gain (NDCG) is obtained by comparing the DCG with the perfect ranking's DCG to get a value between zero and one.

We propose a specific instantiation of the NDCG that matches the AUQC. More specifically, we define an instance's gain $g_i$ as its treatment effect $\tau$. The discount function is a linearly decreasing function: for rank $i$, the discount equals $(n-i+1)$\footnote{Technically, we do not discount lower ranked instances ($d(i) \leq 1$), but rather promote higher ranked instances ($d(i) \geq 1$) \cite[see][eq. 25]{devriendt2020learning}.}. In this specification, we can show that the listwise objective that optimizes the NDCG allows us to learn an optimal ranking policy $\pi$ that optimizes the metric of interest, the AUQC (generalizing \citep[Section 3.2.2]{devriendt2020learning} from the Z-Learner to any model that estimates $\tau$):
\begin{proof}
    The area under the Qini curve (AUQC) is an instantiation of the normalized discounted cumulative gain (NDCG):
    \begin{align*}
        \text{AUQC} 
        = \sum_{k=1}^n \sum_{i=1}^k \tau_{\pi_i}
        = \sum_{i=1}^n \sum_{k=i}^n \tau_{\pi_i}
        = \sum_{i=1}^n \tau_{\pi_i} \sum_{k=i}^n 1
        = \sum_{i=1}^n \tau_{\pi_i} (n - i + 1)
        = \text{NDCG}
    \end{align*}
    for the NDCG with gain $g_{\pi_i} = \tau_{\pi_i}$ and $d(i) = n - i + 1$. 
\end{proof}
Given that the listwise objective described above can be shown to optimize the NDCG \citep{donmez2009local, burges2010lambdamart}, this result proves that our proposed objective can be used to directly optimize the metric of interest: the AUQC. There is one remaining challenge: we do not known the instance's treatment effect $\tau$ and require a valid estimator $\hat{\tau} \rightarrow \tau$. We will tackle this part in \cref{ssec:metalearners}.

\subsubsection{Efficiently scaling to large-scale data sets}
Moving from pointwise to pairwise or listwise optimization requires addressing a challenge regarding computational efficiency. Optimizing over pairs of instances results in an increase of the algorithm's time complexity from $O(n)$ to $O(n^2)$. This complexity is not compatible with the large data sets commonly encountered in applications such as marketing or e-commerce \citep{ai2022lbcf}.

To address this challenge, we propose an efficient sampling procedure that finds a stochastic estimate of the gradient. Intuitively, instead of calculating the gradient based on all possible pairs, we sample $k$ pairs per instance:
\begin{align}
    \mathcal{L_\text{Pair}}(y, \hat{y}) = \sum_{i=1}^n \sum_{j \in \mathcal{J}} - y_{i,j} \log \left( \hat{y}_{i,j} \right) - (1 - y_{i,j}) \big(1 - \log \left( \hat{y}_{i,j} \right)\! \big) \text{ with } \mathcal{J} \sim (\mathcal{U}_{[1, ..., n]})^k, 
    \notag
\end{align}
and equivalently for $\mathcal{L}_\text{List}$. This again makes the procedure scale linearly in the number of instances with complexity $O(kn)$. We observe good results for $k=1$, effectively obtaining the same computational complexity as the pointwise objective. We present a sensitivity analysis for the number of samples $k$ below. We opt for this sampling procedure for its simplicity, although more advanced sampling schemes are possible \citep[see e.g.][]{lyzhin2023tricks}. 

\subsection{Ranking Metalearners}
\label{ssec:metalearners}

One challenge when learning a treatment assignment policy $\pi$ is that decisions need to be made based on the treatment effect $\tau$. Indeed, the optimization of the AUQC presented above requires a model to predict the treatment effect $\hat{\tau}$. Predicting a treatment effect $\tau$ is a challenge. We never observe the treatment effect itself, but only one potential outcome $y(t)$ for each instance, i.e., the outcome when targeted or not targeted--also called the fundamental problem of causal inference \citep{holland1986statistics}. In other words, we never actually observe what we are trying to estimate and optimize over.

To tackle this challenge, we implement the objective functions introduced above for different causal metalearners--general strategies for using any ML method for treatment effect estimation. Whereas metalearners have originally been proposed for effect estimation, we propose adaptations for effect ranking below. In practice, this adaptation consists of integrating ranking (i.e., pairwise or listwise) objectives in each training procedure--instead of the traditional pointwise (regression or classification) objectives. In this section, we describe each metalearner and introduce its ranking equivalent. We focus on several established metalearners, but the extension to other metalearners could be done using a similar approach. While our optimization of the AUQC requires metalearners that directly predict the treatment effect $\tau$, we also discuss adaptations of other metalearners--specifically, the S- and T-Learner.

\paragraph{Z-Learner} The first metalearner estimates a transformation $z$ of the outcome $y$--also called the class transformation approach \cite{jaskowski2012uplift,rudas2018linear,proencca2023incremental}--adjusting the outcome based on the instance's propensity score\footnote{Given that we assume data comes from a randomized trial, we can estimate the propensity score by the proportion of treated instances without fitting a model.} $\hat{e}(x) = P(T=1|X=x)$ and the observed treatment:
\begin{equation*}
    z_{i} =
    \begin{cases}
      y_i / \hat{e}_i   & \text{if $t_i = 1$} \\
      -y_i / (1 - \hat{e}_i)  & \text{if $t_i = 0$}
    \end{cases}
\end{equation*}
The Z-Learner estimates the treatment effect using this outcome: 
\begin{equation*}
    f_\texttt{Z}(x) = \mathbb{E}(Z|X)
    \quad \text{with} \quad
    \hat{\tau} = f_\texttt{Z}(x)
\end{equation*}
Instead of training this final model $f_\texttt{Z}(x)$ with a pointwise objective, we propose to optimize it using a pairwise or listwise objective.

\paragraph{S-Learner} 
The S-Learner estimates a single model, which takes the treatment as a (regular) feature:
\begin{equation*}
    f_\texttt{S}(x, t) = \mathbb{E}(Y|X=x, T=t)
    \quad \text{with} \quad
    \hat{\tau} = f_\texttt{S}(x, 1) - f_\texttt{S}(x, 0)
\end{equation*}
Again, the model $f_\texttt{S}(x, t)$ can be trained with either a pointwise, pairwise, or listwise objective. For the ranking objectives, the estimated treatment effect $\tau$ corresponds to the increase in the ranking score resulting from receiving the treatment. Importantly, however, because this metalearner does not directly optimize for $\tau$, the theoretical results of the previous section do not apply. Nevertheless, although there is no guarantee that the listwise objective optimizes the AUQC for the S-Learner, the difference in ranking scores could provide a good heuristic for the treatment effect.

\paragraph{T-Learner}
The T-Learner trains two models: one model for each treatment group--- $f_{\texttt{T}_1}$ for the treatment ($T=1$) and $f_{\texttt{T}_0}$ for the control group ($T=0$)---trained as follows:
\begin{equation*}
    f_{\texttt{T}_1}(x) = \mathbb{E}(Y|X=x, T=1),
    \quad 
    f_{\texttt{T}_0}(x) = \mathbb{E}(Y|X=x, T=0).
\end{equation*}
When combined, these can estimate the treatment effect as follows:
\begin{equation*}
    \hat{\tau} = f_{\texttt{T}_1}(x) - f_{\texttt{T}_0}(x).
\end{equation*}
We propose to train both models $f_{\texttt{T}_1}(x)$ and $f_{\texttt{T}_0}(x)$ with a pairwise or listwise objective, instead of the traditional pointwise objective. This corresponds to a separate optimization of the AUQCof the treatment ($f_{\texttt{T}_1}(x)$) and control ($f_{\texttt{T}_0}(x)$) groups (based on the outcome $y$ instead of the effect $\tau$). An instance's difference in ranking scores between both groups is used as a proxy for its treatment effect. Similarly to the S-Learner, the theoretical results from the previous section do not apply. Nevertheless, this separate optimization could be a good heuristic for the AUQC \citep[see also][eq. 26]{devriendt2018literature}.

\paragraph{X-Learner} The X-Learner first estimates an initial treatment effect by imputing the counterfactual potential outcome using a T-Learner model as follows \cite{kunzel2019metalearners}:
\begin{equation*}
      D^1_i = y_i - f_{\texttt{T}_0}(x_i)  \,\, \text{if $t_i = 1$}, \quad
      D^0_i = f_{\texttt{T}_1}(x_i) - y_i  \,\, \text{if $t_i = 0$}.
\end{equation*}
The final two models are then trained on the imputed effects:
\begin{equation*}
    \hat{\tau}^0 = f^0_\texttt{X}(x) = \mathbb{E}(D^0_i|X=x), 
    \quad \hat{\tau}^1 = f^1_\texttt{X}(x) = \mathbb{E}(D^1_i|X=x). 
\end{equation*}
To obtain the final predicted effect $\hat{\tau}_i$, we combine these two models:
\begin{equation*}
    \quad \text{with} \quad
    \hat{\tau} = g(x) f^0_X(x) + (1 - g(x)) f^1_X(x). 
\end{equation*}
For the weighting function $g(x)$, we use the estimated propensity score $\hat{e}(x) = P(T|X=x)$ following \cite{kunzel2019metalearners}. Compared to the pointwise variant, we propose to train the final models $f^0_\texttt{X}(x)$ and $f^1_\texttt{X}(x)$ with pairwise or listwise ranking objectives, with the initial models still trained using a pointwise objective. The final ranking score $\hat{\tau}$ is a linear combination of two ranking scores \citep[see][note 7.1]{burges2010lambdamart}. 

\paragraph{DR-Learner}
The Doubly Robust or DR-Learner \citep{bang2005doubly,kennedy2023towards} also relies on a final model estimating a pseudo-outcome. In this case, the first stage is based on pointwise estimates from a T-Learner and a propensity model $\hat{e}(x) = P(T|X=x)$. These estimates are combined to create a pseudo-outcome $\phi$ as follows:
\begin{equation*}
    \phi_i = \frac{t_i - \hat{e}(x_i)}{\hat{e}(x_i) (1 - \hat{e}(x_i))} (y_i - f_{\texttt{T}_{t_i}}(x_i)) + f_{\texttt{T}_1}(x_i) - f_{\texttt{T}_0}(x_i).
\end{equation*}
This pseudo-outcome is then used to learn a final model $\hat{\tau} = f_\texttt{DR}(\phi | x)$, which can be learned using a point-, pair-, or listwise objective.

\paragraph{R-Learner}
For the R-Learner \citep{nie2021quasi}, we first fit an outcome model $\hat{m}(x) = \mathbb{E}(Y|X=x)$ and propensity model $\hat{e}(x) = P(T|X=x)$. These can then be used to minimize the R-Loss, based on Robinson's decomposition \cite{robinson1988root}, which can be seen as a weighted MSE: 
\begin{equation*}
\begin{split}
    \mathcal{L}^R_\text{MSE}(y, \hat{y}) = \frac{1}{n} \sum_{i=1}^n \left( \left( y_i - \hat{m}(x_i) \right) - \left( t_i - \hat{e}(x_i) \right) \tau (x_i) \right)^2
    = \frac{1}{n} \sum_{i=1}^n \frac{1}{\left( t_i - \hat{e}(x_i) \right)^2} \left( \left( \frac{y_i - \hat{m}(x_i)}{t_i - \hat{e}(x_i)}\right)  - \tau (x_i) \right)^2
    .
\end{split}
\end{equation*}
Instead of this pointwise objective, we propose to use a weighted pair- or listwise objective based on the same weights and labels. 

For simplicity, we do not use out-of-fold estimates for any of the intermediary models for any of the metalearners, but rather train all models on the same train set.

\section{Empirical Results}
\label{sec:empirical_results}

\begin{figure*}[t]
\centering
\begin{subfigure}[b]{0.33\textwidth}
    \centering
    \includegraphics[height=0.45\linewidth]{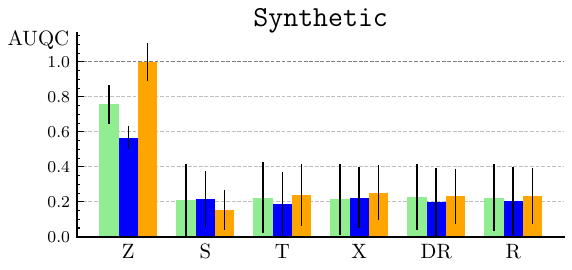}
\end{subfigure}
\hfill
\begin{subfigure}[b]{0.33\textwidth}
    \centering
    \includegraphics[height=0.45\linewidth]{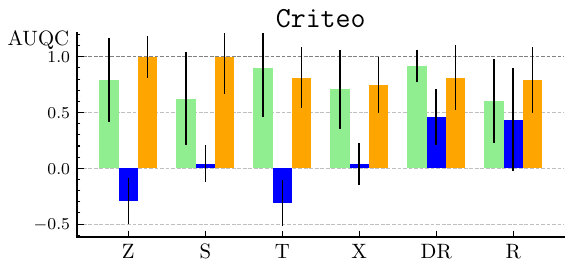}
\end{subfigure}
\hfill
\begin{subfigure}[b]{0.33\textwidth}
    \centering
    \includegraphics[height=0.45\linewidth]{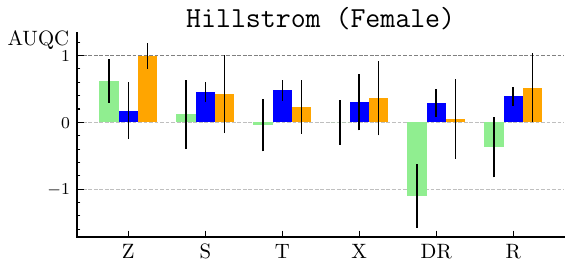}
\end{subfigure}

% \par\bigskip
\vspace{2pt}

\hspace{5pt}
\begin{subfigure}[c]{0.33\textwidth}
    \centering
    \includegraphics[height=0.45\linewidth]{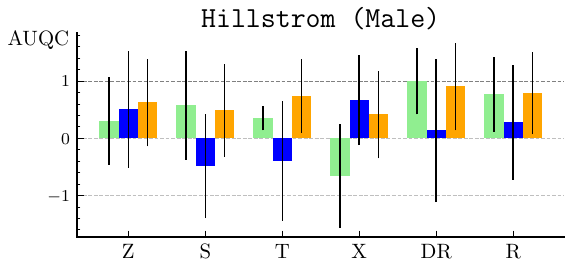}
\end{subfigure}%
\hspace{5pt}
\begin{subfigure}[c]{0.33\textwidth}
    \centering
    \includegraphics[height=0.45\linewidth]{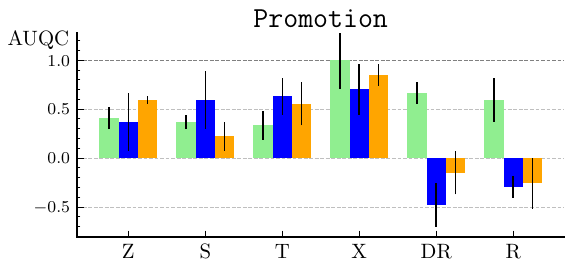}
\end{subfigure}%
\hspace{25pt}
\begin{subfigure}[c]{0.1\textwidth}
    \centering
    \includegraphics[width=\linewidth]{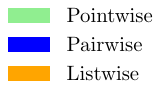}
\end{subfigure}
\hspace{5pt}

\vspace{-8pt}
\caption{\textit{Ranking Quality for Different Objectives and Metalearners.} For each metalearner, we compare a point-, pair-, and listwise version. We show the AUQC $\pm$ one standard error, scaled to have the best result $= 1$, for five different data sets.
}
\vspace{-5pt}
\noindent\rule{\textwidth}{0.1pt}
\vspace{-16pt}
\label{fig:results_overview_overall}
\end{figure*}

This section presents the empirical results, comparing our proposed (pairwise and listwise) \textit{effect ranking} metalearners with traditional (pointwise) \textit{effect estimation} metalearners. Our experiments aim to answer three research questions. (\textit{RQ1}) What is the treatment recommendation quality resulting from the different methods, as measured in AUQC? (\textit{RQ2}) What are the performance trade-offs of the pointwise, pairwise, and listwise objectives, in terms of MSE (i.e., pointwise accuracy), Kendall $\tau$ (i.e., pairwise rank correlation), and AUQC (i.e., listwise ranking quality)? (\textit{RQ3}) How sensitive are our proposed methods to key hyperparameters? This section presents the setup of our experimental evaluation and the empirical results.

% \subsection{Data and evaluation metrics}

\subsection{Data and Benchmarks}
To evaluate the performance of the proposed approaches, we use a total of four data sets based on randomized trials: 1) a synthetic dataset; 2) Criteo \cite{diemert2018large}; 3) Hillstrom (Male and Female) \cite{hillstrom2008minethatdata}; 4) a proprietary data set from a promotion campaign at a global online travel agency.

We first simulate a \texttt{Synthetic} data set. Simulated data allows for a more comprehensive evaluation than real data, as we know the treatment effect for test instances. 
% Some references
The data generating process is inspired by an e-commerce setting and similar to existing work \cite{proencca2023incremental}. First, we generate customer characteristics as follows: $X \sim \mathcal{N}(0, 1)^d$. Then, we generate a sale (or conversion) probability $S$ based on these characteristics and random coefficients $U_s \sim \mathcal{U}(-1, 1)^d$ as $S = \frac{1}{1 + \exp{\left(- \sum_d U_s X + \epsilon_s\right)}}$, where $\epsilon_s \sim \mathcal{N}(0, 0.1)$. Similarly, we generate a potential revenue $R$ using random coefficients $U_r \sim \mathcal{U}(-1, 1)^d$ as $R = 1 + \left| \sum_d U_r X \right | + \epsilon_r$, where $\epsilon_r \sim \mathcal{N}(0, 0.1)$. The cost of the treatment $C$ (the marketing incentive or discount) is defined as $10\%$ of the revenue: $C = 0.1 R$. The observed outcome $Y$, the net revenue generated for that customer, is the revenue for that customer minus the treatment cost--simulated as follows:
\begin{equation*}
    y_{i} =
    \begin{cases}
        r_i - c_i   & \text{if $t_i = 1$ and $s_i=1$} \\
        -c_i        & \text{if $t_i = 1$ and $s_i=0$} \\
        r_i         & \text{if $t_i = 0$ and $s_i=1$} \\
        0           & \text{if $t_i = 0$ and $s_i=0$}.
    \end{cases}
\end{equation*}
We generate $10,000$ instances with $d=10$ characteristics. 

Next, we also compare with three real-world data sets. The \texttt{Criteo} data set \citep{diemert2018large} is the result of a randomized trial testing whether showing an advertisement increases a customer's visit or conversion probability. For the outcome, we follow \cite{zhou2023direct} and take the net revenue $y$ as conversion minus visit. To reduce training times, we randomly sample $500,000$ instances from this data set.

The \texttt{Hillstrom} data set was collected to test whether an e-mail campaign resulted in additional sales. Two treatments were recorded: a \texttt{Men}'s and \texttt{Women}'s e-mail. Therefore, we split the data in two data sets for both treatments, and use the same control group for both. We calculate each customer's net revenue as revenue (conversion times spend) minus visit.

Finally, data from a \texttt{Promotion} campaign at a large online travel agency was used as a randomized dataset to evaluate the offline performance of different approaches in a real-world setting.

We compare the three objectives and metalearners for gradient boosting based on LightGBM \cite{ke2017lightgbm}. We implement our proposed pairwise and listwise objectives as a custom objectives and metrics in LightGBM.  
For each data set, we run a five-fold cross-validation procedure. We run hyperparameter tuning using 10 random sampling iterations over the following hyperparameters: ``\texttt{num\_leaves}'' $\in [10, 50]$, ``\texttt{learning\_rate}'' $\in [0.01, 0.20]$, ``\texttt{max\_depth}'' $\in [3, 10]$, and ``\texttt{min\_data\_in\_leaf}'' $\in [10, 30]$. We use $64\%$ of all instances for training, $16\%$ for validation, and $20\%$ for testing. 

\begin{figure*}[t]
\centering
    \begin{subfigure}[b]{0.32\textwidth}
        \centering
        \includegraphics[width=\textwidth]{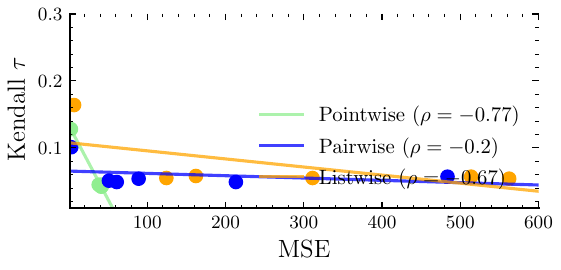}
        % \caption{MSE vs Kendall $\tau$}
    \end{subfigure}
    \hfill
    \begin{subfigure}[b]{0.32\textwidth}
        \centering
        \includegraphics[width=\textwidth]{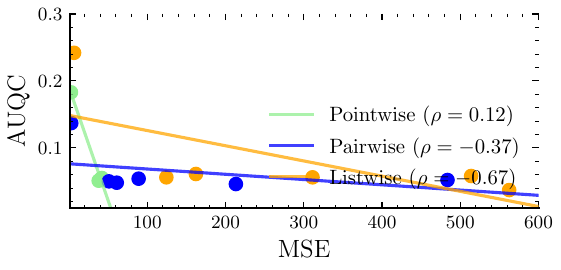}
        % \caption{MSE vs AUQC}
    \end{subfigure}
    \hfill
    \begin{subfigure}[b]{0.32\textwidth}
        \centering
        \includegraphics[width=\textwidth]{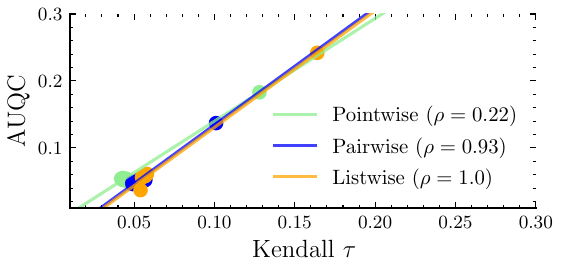}
        % \caption{Kendall $\tau$ vs AUQC}
    \end{subfigure}
% \vspace{-10pt}
\caption{\textit{Analyzing Performance Trade-offs on Synthetic Data.} We compare the three different objectives (point-, pair-, and listwise) across metalearners. Using the \texttt{Synthetic} data set, we compare performance in terms of MSE (i.e., pointwise accuracy), Kendall $\tau$ (i.e., pairwise rank correlation), and AUQC (i.e., listwise decision quality). For each, we show the correlation $\rho$.}
% \noindent\rule{\textwidth}{0.1pt}
\label{fig:results_overview_synthetic}
\end{figure*}

% RQ3: Sampling iterations
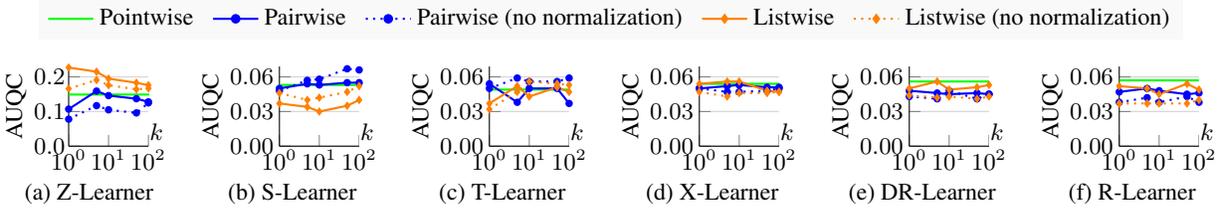
\begin{figure*}
    \centering
    \begin{subfigure}[b]{\linewidth}
    \centering
        \begin{tikzpicture}[]%
        \begin{axis}[
          hide axis, 
          axis lines=middle,
          xmin=1,
          xmax=1000,
          ymin=0.075,
          ymax=0.325,
          width=\linewidth,
          height=\linewidth,
          legend style={
              draw=none,
              fill=gray!5, 
              font=\small,
              /tikz/every even column/.append style={column sep=5pt},
          },
          legend cell align={left},
          legend columns=5,
          every axis plot/.append style={thick},
        ]
        
        \addlegendimage{color=green, mark=none}
        \addlegendentry{Pointwise}
        
        \addlegendimage{color=blue, mark=*, mark options={scale=0.7, solid}}
        \addlegendentry{Pairwise}
        
        \addlegendimage{color=blue, mark=*, dotted, mark options={scale=0.7, solid}}
        \addlegendentry{Pairwise (\text{no normalization})}
        
        \addlegendimage{color=orange, mark=diamond*, mark options={scale=0.7, solid}}
        \addlegendentry{Listwise}
        
        \addlegendimage{color=orange, mark=diamond*, dotted, mark options={scale=0.7, solid}}
        \addlegendentry{Listwise (\text{no normalization})}
        \end{axis}
        \end{tikzpicture}
    \end{subfigure}
    \vspace{-4pt}
    
    \begin{subfigure}[b]{0.16\linewidth}
        \centering
        \begin{tikzpicture}
        \begin{axis}[
          axis lines*=middle,
          ymajorgrids,
          grid style={line width=.1pt, draw=gray!10},
          major grid style={line width=0.5pt,draw=gray!30},
          xmin=1,
          xmax=100,
          xmode=log,
          ymin=0.0,
          ymax=0.23,
          xlabel=$k$,
          ylabel={\small AUQC},
          xlabel style={at={(1.1, 0.4)}, font=
        \small},
          ylabel style={rotate=-90, font=
        \small},
          ylabel shift = -4pt,
          ylabel near ticks, 
          xtick={1, 10, 100, 1000},
          ytick={0.0, 0.1, ..., 0.4},
          hide obscured x ticks=false,
          hide obscured y ticks=false,
          width=\linewidth,
          height=\linewidth,
          every axis plot/.append style={thick},
          y tick label style={
            /pgf/number format/.cd,
            fixed,
            fixed zerofill,
            precision=1,
            /tikz/.cd,
            font=\footnotesize, 
            xshift=0.5ex,
          },
          xminorticks=false,
          xticklabel style = {font=\footnotesize, yshift=0.5ex},
        ]
        \addplot[color=green, mark=none, error bars/.cd, y dir=both, y explicit]
            coordinates {
            (1, 0.149 )
            (1000, 0.149 )
            };
        \addplot[color=blue, mark=*, mark options={scale=0.5, solid}]
            coordinates {
            (1, 0.107 )
            (5, 0.159 )
            (10, 0.146 )
            (50, 0.137 )
            (100, 0.128 )
            };
        \addplot[color=blue, mark=*, dotted, mark options={scale=0.5, solid}]
            coordinates {
            (1, 0.078 )
            (5, 0.117 )
            (10, 0.105 )
            (50, 0.096 )
            (100, 0.125 )
            };
        \addplot[color=orange, mark=diamond*, mark options={scale=0.5, solid}]
            coordinates {
            (1, 0.227 )
            (5, 0.215 )
            (10, 0.195 )
            (50, 0.183 )
            (100, 0.176 )
            };
        \addplot[color=orange, mark=diamond*, dotted, mark options={scale=0.5, solid}]
            coordinates {
            (1, 0.166 )
            (5, 0.191 )
            (10, 0.176 )
            (50, 0.164 )
            (100, 0.167 )
            };
        \end{axis}
        \end{tikzpicture}
        \vspace{-5pt}
        \caption{Z-Learner}
    \end{subfigure}%
    \hfill
    \begin{subfigure}[b]{0.16\textwidth}
        \centering
        \begin{tikzpicture}
        \begin{axis}[
          axis lines*=middle,
          ymajorgrids,
          grid style={line width=.1pt, draw=gray!10},
          major grid style={line width=0.5pt,draw=gray!30},
          xmin=1,
          xmax=100,
          xmode=log,
          ymin=0.0,
          ymax=0.069,
          xlabel=$k$,
          ylabel={\small AUQC},
          xlabel style={at={(1.1, 0.4)}, font=
        \small},
          ylabel style={rotate=-90, font=
        \small},
          ylabel shift = -4pt,
          ylabel near ticks, 
          xtick={1, 10, 100, 1000},
          ytick={0.0, 0.03, ..., 0.4},
          hide obscured x ticks=false,
          hide obscured y ticks=false,
          width=\linewidth,
          height=\linewidth,
          every axis plot/.append style={thick},
          y tick label style={
            /pgf/number format/.cd,
            fixed,
            fixed zerofill,
            precision=2,
            /tikz/.cd,
            font=\footnotesize, 
            xshift=0.3ex,
          },
          xminorticks=false,
          xticklabel style = {font=\footnotesize, yshift=0.5ex},
        ]
        \addplot[color=green, mark=none, ]
            coordinates {
            (1, 0.053 )
            (1000, 0.053 )
            };
        \addplot[color=blue, mark=*, mark options={scale=0.5, solid}]
            coordinates {
            (1, 0.050 )
            (5, 0.054 )
            (10, 0.053 )
            (50, 0.055 )
            (100, 0.055 )
            };
        \addplot[color=blue, mark=*, dotted, mark options={scale=0.5, solid}]
            coordinates {
            (1, 0.048 )
            (5, 0.057 )
            (10, 0.058 )
            (50, 0.067 )
            (100, 0.066 )
            };
        \addplot[color=orange, mark=diamond*, mark options={scale=0.5, solid}]
            coordinates {
            (1, 0.037 )
            (5, 0.034 )
            (10, 0.030 )
            (50, 0.035 )
            (100, 0.040 )
            };
        \addplot[color=orange, mark=diamond*, dotted, mark options={scale=0.5, solid}]
            coordinates {
            (1, 0.046 )
            (5, 0.040 )
            (10, 0.042 )
            (50, 0.047 )
            (100, 0.052 )
            };
        \end{axis}
        \end{tikzpicture}
        \vspace{-5pt}
        \caption{S-Learner}
    \end{subfigure}
    \hfill
    \begin{subfigure}[b]{0.16\textwidth}
    \centering
        \begin{tikzpicture}
        \begin{axis}[
          axis lines*=middle,
          ymajorgrids,
          grid style={line width=.1pt, draw=gray!10},
          major grid style={line width=0.5pt,draw=gray!30},
          xmin=1,
          xmax=100,
          xmode=log,
          ymin=0.0,
          ymax=0.069,
          xlabel=$k$,
          ylabel={\small AUQC},
          xlabel style={at={(1.1, 0.4)}, font=
        \small},
          ylabel style={rotate=-90, font=
        \small},
          ylabel shift = -4pt,
          ylabel near ticks, 
          xtick={1, 10, 100, 1000},
          ytick={0.0, 0.03, ..., 0.4},
          hide obscured x ticks=false,
          hide obscured y ticks=false,
          width=\linewidth,
          height=\linewidth,
          every axis plot/.append style={thick},
          y tick label style={
            /pgf/number format/.cd,
            fixed,
            fixed zerofill,
            precision=2,
            /tikz/.cd,
            font=\footnotesize, 
            xshift=0.3ex,
          },
          xminorticks=false,
          xticklabel style = {font=\footnotesize, yshift=0.5ex},
        ]
        \addplot[color=green, mark=none, ]
            coordinates {
            (1, 0.049 )
            (1000, 0.049 )
            };
        \addplot[color=blue, mark=*, mark options={scale=0.5, solid}]
            coordinates {
            (1, 0.054 )
            (5, 0.038 )
            (10, 0.050 )
            (50, 0.050 )
            (100, 0.037 )
            };
        \addplot[color=blue, mark=*, dotted, mark options={scale=0.5, solid}]
            coordinates {
            (1, 0.050 )
            (5, 0.059 )
            (10, 0.056 )
            (50, 0.056 )
            (100, 0.059 )
            };
        \addplot[color=orange, mark=diamond*, mark options={scale=0.5, solid}]
            coordinates {
            (1, 0.037 )
            (5, 0.052 )
            (10, 0.043 )
            (50, 0.051 )
            (100, 0.047 )
            };
        \addplot[color=orange, mark=diamond*, dotted, mark options={scale=0.5, solid}]
            coordinates {
            (1, 0.032 )
            (5, 0.047 )
            (10, 0.056 )
            (50, 0.054 )
            (100, 0.053 )
            };
        \end{axis}
        \end{tikzpicture}
        \vspace{-5pt}
        \caption{T-Learner}
    \end{subfigure}
    \hfill    
    \begin{subfigure}[b]{0.16\linewidth}
        \centering
        \begin{tikzpicture}
        \begin{axis}[
          axis lines*=middle,
          ymajorgrids,
          grid style={line width=.1pt, draw=gray!10},
          major grid style={line width=0.5pt,draw=gray!30},
          xmin=1,
          xmax=100,
          xmode=log,
          ymin=0.0,
          ymax=0.069,
          xlabel=$k$,
          ylabel={\small AUQC},
          xlabel style={at={(1.1, 0.4)}, font=
        \small},
          ylabel style={rotate=-90, font=
        \small},
          ylabel shift = -4pt,
          ylabel near ticks, 
          xtick={1, 10, 100, 1000},
          ytick={0.0, 0.03, ..., 0.4},
          hide obscured x ticks=false,
          hide obscured y ticks=false,
          width=\linewidth,
          height=\linewidth,
          every axis plot/.append style={thick},
          y tick label style={
            /pgf/number format/.cd,
            fixed,
            fixed zerofill,
            precision=2,
            /tikz/.cd,
            font=\footnotesize, 
            xshift=0.3ex,
          },
          xminorticks=false,
          xticklabel style = {font=\footnotesize, yshift=0.5ex},
        ]
        \addplot[color=green, mark=none, ]
            coordinates {
            (1, 0.054 )
            (1000, 0.054 )
            };
        \addplot[color=blue, mark=*, mark options={scale=0.5, solid}]
            coordinates {
            (1, 0.050 )
            (5, 0.052 )
            (10, 0.053 )
            (50, 0.051 )
            (100, 0.051 )
            };
        \addplot[color=blue, mark=*, dotted, mark options={scale=0.5, solid}]
            coordinates {
            (1, 0.050 )
            (5, 0.047 )
            (10, 0.047 )
            (50, 0.048 )
            (100, 0.050 )
            };
        \addplot[color=orange, mark=diamond*, mark options={scale=0.5, solid}]
            coordinates {
            (1, 0.054 )
            (5, 0.056 )
            (10, 0.056 )
            (50, 0.048 )
            (100, 0.047 )
            };
        \addplot[color=orange, mark=diamond*, dotted, mark options={scale=0.5, solid}]
            coordinates {
            (1, 0.047 )
            (5, 0.043 )
            (10, 0.046 )
            (50, 0.046 )
            (100, 0.047 )
            };
        \end{axis}
        \end{tikzpicture}
        \vspace{-5pt}
        \caption{X-Learner}
    \end{subfigure}
    \hfill
    \begin{subfigure}[b]{0.16\textwidth}
        \centering
        \begin{tikzpicture}
        \begin{axis}[
          axis lines*=middle,
          ymajorgrids,
          grid style={line width=.1pt, draw=gray!10},
          major grid style={line width=0.5pt,draw=gray!30},
          xmin=1,
          xmax=100,
          xmode=log,
          ymin=0.0,
          ymax=0.069,
          xlabel=$k$,
          ylabel={\small AUQC},
          xlabel style={at={(1.1, 0.4)}, font=
        \small},
          ylabel style={rotate=-90, font=
        \small},
          ylabel shift = -4pt,
          ylabel near ticks, 
          xtick={1, 10, 100, 1000},
          ytick={0.0, 0.03, ..., 0.4},
          hide obscured x ticks=false,
          hide obscured y ticks=false,
          width=\linewidth,
          height=\linewidth,
          every axis plot/.append style={thick},
          y tick label style={
            /pgf/number format/.cd,
            fixed,
            fixed zerofill,
            precision=2,
            /tikz/.cd,
            font=\footnotesize, 
            xshift=0.3ex,
          },
          xminorticks=false,
          xticklabel style = {font=\footnotesize, yshift=0.5ex},
        ]
        \addplot[color=green, mark=none, ]
            coordinates {
            (1, 0.056 )
            (1000, 0.056 )
            };
        \addplot[color=blue, mark=*, mark options={scale=0.5, solid}]
            coordinates {
            (1, 0.048 )
            (5, 0.046 )
            (10, 0.045 )
            (50, 0.046 )
            (100, 0.045 )
            };
        \addplot[color=blue, mark=*, dotted, mark options={scale=0.5, solid}]
            coordinates {
            (1, 0.043 )
            (5, 0.041 )
            (10, 0.048 )
            (50, 0.041 )
            (100, 0.045 )
            };
        \addplot[color=orange, mark=diamond*, mark options={scale=0.5, solid}]
            coordinates {
            (1, 0.050 )
            (5, 0.056 )
            (10, 0.049 )
            (50, 0.051 )
            (100, 0.053 )
            };
        \addplot[color=orange, mark=diamond*, dotted, mark options={scale=0.5, solid}]
            coordinates {
            (1, 0.044 )
            (5, 0.042 )
            (10, 0.043 )
            (50, 0.041 )
            (100, 0.043 )
            };
        \end{axis}
        \end{tikzpicture}
        \vspace{-5pt}
        \caption{DR-Learner}
    \end{subfigure}
    \hfill
    \begin{subfigure}[b]{0.16\textwidth}
    \centering
        \begin{tikzpicture}
        \begin{axis}[
          axis lines*=middle,
          ymajorgrids,
          grid style={line width=.1pt, draw=gray!10},
          major grid style={line width=0.5pt,draw=gray!30},
          xmin=1,
          xmax=100,
          xmode=log,
          ymin=0.0,
          ymax=0.069,
          xlabel=$k$,
          ylabel={\small AUQC},
          xlabel style={at={(1.1, 0.4)}, font=
        \small},
          ylabel style={rotate=-90, font=
        \small},
          ylabel shift = -4pt,
          ylabel near ticks, 
          xtick={1, 10, 100, 1000},
          ytick={0.0, 0.03, ..., 0.4},
          hide obscured x ticks=false,
          hide obscured y ticks=false,
          width=\linewidth,
          height=\linewidth,
          every axis plot/.append style={thick},
          y tick label style={
            /pgf/number format/.cd,
            fixed,
            fixed zerofill,
            precision=2,
            /tikz/.cd,
            font=\footnotesize, 
            xshift=0.3ex,
          },
          xminorticks=false,
          xticklabel style = {font=\footnotesize, yshift=0.5ex},
        ]
        \addplot[color=green, mark=none, ]
            coordinates {
            (1, 0.057 )
            (1000, 0.057 )
            };
        \addplot[color=blue, mark=*, mark options={scale=0.5, solid}]
            coordinates {
            (1, 0.047 )
            (5, 0.050 )
            (10, 0.048 )
            (50, 0.045 )
            (100, 0.046 )
            };
        \addplot[color=blue, mark=*, dotted, mark options={scale=0.5, solid}]
            coordinates {
            (1, 0.038 )
            (5, 0.042 )
            (10, 0.038 )
            (50, 0.043 )
            (100, 0.038 )
            };
        \addplot[color=orange, mark=diamond*, mark options={scale=0.5, solid}]
            coordinates {
            (1, 0.052 )
            (5, 0.050 )
            (10, 0.045 )
            (50, 0.054 )
            (100, 0.049 )
            };
        \addplot[color=orange, mark=diamond*, dotted, mark options={scale=0.5, solid}]
            coordinates {
            (1, 0.037 )
            (5, 0.037 )
            (10, 0.037 )
            (50, 0.037 )
            (100, 0.040 )
            };
        \end{axis}
        \end{tikzpicture}
        \vspace{-5pt}
        \caption{R-Learner}
    \end{subfigure}
% \vspace{-8pt}
\caption{\textit{What Is the Effect of the Number of Sampling Iterations $k$?} We show performance in terms of AUQC for the different metalearners on the \texttt{Synthetic} data set. We fix the sigmoid parameter $\sigma = 1$ and train with default hyperparameters.}
% \vspace{-5pt}
\noindent\rule{\textwidth}{0.1pt}
\vspace{-16pt}
\label{fig:experiment_synthetic_sampling_iterations_main}
\end{figure*}

% Main results
\subsection{Comparing Performance for the Different Objectives and Metalearners {(RQ1)}}

% Overall performance
For each metalearner, we compare a model trained with only a pointwise objective to our proposed ranking alternatives, based on either pairwise and listwise objectives. We evaluate the quality of each treatment allocation policy by looking at the cumulative treatment effect over the instance ranking, measured using the AUQC presented above. We evaluate the performance for each metalearner and objective over the different data sets (see \cref{fig:results_overview_overall} and \cref{ssec:rq1_additional}). Across data sets and metalearners, we observe that listwise metalearners generally result in better treatment prioritization.
Over all tested data sets and metalearners, a pointwise objective gives the highest AUQC in only a minority of cases (7/30), while the listwise objective obtains best in class performance in a majority of time (16/30) and the pairwise objective performs similar to the pointwise (7/30). A listwise approach outperforms a pointwise one in a majority of cases (20/30).
Only for the \texttt{Promotion} data, the pointwise objective performs relatively well. 

Interestingly, we observe differences across metalearners in terms of which objective gives the best results. The listwise objective seems favorable for some metalearners (Z-, X-, and R-Learner), while the pairwise objective seems preferable for the S-Learner and the pointwise objective seems best for the DR-Learner.
Generally, we also observe that the choice of metalearner is at least equally important as the choice of objective. This finding stresses the importance of testing different metalearners--illustrating the value of our contribution.

When learning a model using ranking objectives, the ranking scores are not properly calibrated. This represents a possible challenge for metalearners that do not directly learn the treatment effect, i.e., the S- and T-Learner. For these metalearners, an instance's predicted ranking score does not necessarily correspond to its potential outcome. Rather, if one instance's outcome is larger than another instance's ($y_i > y_j$), its predicted score will be larger ($\hat{y}_i > \hat{y}_j$). Although the ranking might hold for both potential outcomes, there are no guarantees for the rankings of the estimated treatment effects derived from these ranking scores ($\tau_i$ and $\tau_j$). Because ranking scores are not calibrated, arithmetic operations of the scores (as used by these two metalearners) may not be meaningful. Nevertheless, the ranking versions of the S- and T-Learner perform relatively well in practice, illustrating that they may provide a meaningful heuristic. While calibration methods for ranking models exists, we leave this extension for future work (see conclusion).

\subsection{Analyzing Alternative Metrics {(RQ2)} and Design Choices {(RQ3)}}

This section aims to provide a deeper understanding of our proposed approach. To this end, we highlight additional results using the \texttt{Synthetic} data, which allows for a more comprehensive analysis as we know the ground truth treatment effects for the test instances. 

To answer \textit{(RQ2)} regarding performance trade-offs of the different objectives and metalearners, we can present additional metrics to allow for a more holistic evaluation (see \cref{fig:results_overview_synthetic} and \cref{ssec:rq2_3_additional}). Our main metric of interest remains the AUQC: a listwise metric of ranking quality. Additionally, we present two other metrics: a pointwise error metric (MSE) and a pairwise rank correlation coefficient (Kendall $\tau$). First, we observe that ranking objectives give far worse performance in terms of MSE (\cref{ssec:rq2_3_additional}). In other words, the ranking score does not accurately reflect the size of the treatment effect. In relation to this finding, \cref{fig:results_overview_synthetic} shows that the MSE is not a good predictor of performance in terms of AUQC. Conversely, Kendall $\tau$ is a good predictor for AUQC for all models ($\rho > 0$), particularly for the ranking models ($\rho > 0.9$). This finding underscores the importance of ranking metrics for evaluating decision quality and highlights the irrelevance of optimizing pointwise error for treatment prioritization.

Finally, to answer \textit{(RQ3)}, we analyze several design choices of our proposed ranking metalearners in \cref{fig:experiment_synthetic_sampling_iterations_main} and \cref{ssec:sensitivity_analysis}. We observe that the default setting used in the experiments above (sampling iterations $k=1$, sigmoid $\sigma=1$, normalizing ranking scores) generally performs well across ranking objectives and metalearners. Interestingly, we observe little performance benefits when training with more sampling iterations. This finding shows that our sampled objective can estimate the ranking objective accurately. Only for the S- and T-Learner without normalization, we observe better performance for a higher $k$. Additionally, the stochasticity of our sampled objectives may even provide a form of regularization \citep{friedman2002stochastic}. 

\section{Conclusion}

% What we propose
This work addressed the problem of optimally prioritizing instances for treatment, an important problem for many applications where not all instances can receive a treatment. Existing approaches typically tackle this problem in a prediction-focused approach by first obtaining a pointwise \textit{effect estimate} for each instance's treatment effect, and then ranking instances based on these estimates. Conversely, we explore an alternative, decision-focused approach: using objectives that learn to \textit{rank the treatment effects}, we aim to optimize the quality of the resulting treatment policy directly. Building on the literature on learning to rank, we propose pairwise and listwise ranking objectives and show that our proposed listwise objective directly optimizes the policy's AUQC. Moreover, we propose different ranking metalearners by integrating these ranking objectives in the construction of each metalearner. Empirical results show that our proposed \textit{effect ranking} approach can outperform a pointwise, \textit{effect estimation} approach. In conclusion, our proposed ranking metalearners offer a valuable tool for applications where instances need to be prioritized for treatment.

% Limitations and future work
Our work opens up several exciting directions for future work. First, by building upon advances in learning to rank, such as more advanced listwise objectives \cite[e.g.,][]{lyzhin2023tricks} or calibration of ranking scores \cite[e.g.,][]{sculley2010combined, yan2022scale}. Alternatively, we envision extensions of our approach to more complicated settings with, e.g., more advanced operational constraints \citep{vanderschueren2022new}, multiple or continuous treatments, or more complex objectives (e.g., incremental return on investment). Additionally, it would be insightful to analyze how our proposed ranking metalearners perform when learning from confounded observational data with non-random treatment assignments. Finally, while theoretical results regarding convergence or error bounds are out of the scope of this work, we believe that extending the results obtainedfor the effect estimation models \citep[e.g.,][]{nie2021quasi,kennedy2023towards} to effect ranking models is a fruitful area of future work.

\bibliographystyle{unsrt}
\bibliography{references}

\clearpage
\appendix

\section{Problem formulation: Identifiability Assumptions}
\label{sec:assumptions_app}

As introduced in the main body, we require the standard assumptions from causal inference to identify the causal effect. In this work, we assumed that historical data comes from a randomized controlled trial: 
\begin{assumption}[Consistency]
    When $Y=y$ and $T=t$, we assume that $Y(T=t) = y$. This implicates that, for each instance, when given treatment $t$, the outcome we observe is the potential outcome associated to that treatment $Y(t)$.
\end{assumption}
\begin{assumption}[No interference]
    An instance's outcome given a treatment is independent of treatments administered to other instances: $Y_i(t_0, \dots, t_i, \dots, t_n) = Y_i(t_i)$.
\end{assumption}
\begin{assumption}[Unconfoundedness]
    We assume $Y(T) \ind T$, i.e., past treatment decisions were made at random, i.e., not based on the instance's characteristics.
\end{assumption}
If we do not have data from a randomized trial, we require a stronger assumption called strong ignorability or no hidden confounding: $Y(T) \ind T | x$, i.e., past treatment decisions were exclusively based on the instance's observed characteristics $x$. In this case, we also require positivity: for each instance, the probability of administering each treatment has to be larger than zero, i.e., $P(T | x) > 0$. We do not consider this scenario in our work. However, our ranking metalearners can easily be extended to these scenarios: while some metalearners already integrate the propensity score in their construction, others may be improved by inverse propensity score weighting \citep{horvitz1952generalization,hernan2006estimating}.

\section{Empirical Results: Additional Experiments}

This section presents additional experimental results. We describe additional results for RQ1 in \cref{ssec:rq1_additional}, and for RQ2 and RQ3 in \cref{ssec:rq2_3_additional} where we analyze the effect of several design choices and hyperparameters.

\subsection{Comparing Performance for the Different Objectives and Metalearners (RQ1): Additional Results}
\label{ssec:rq1_additional}

\begin{figure*}[t]
\centering
\hspace{5pt}
\begin{subfigure}[b]{0.33\textwidth}
    \centering
    \includegraphics[height=0.45\linewidth]{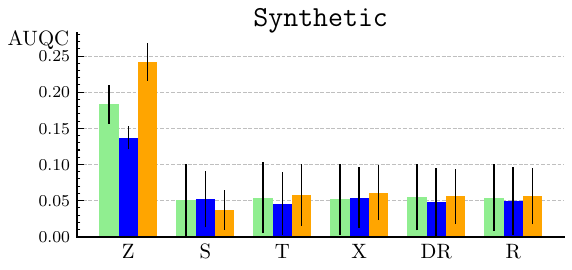}
\end{subfigure}
\hspace{15pt}
\begin{subfigure}[b]{0.33\textwidth}
    \centering
    \includegraphics[height=0.45\linewidth]{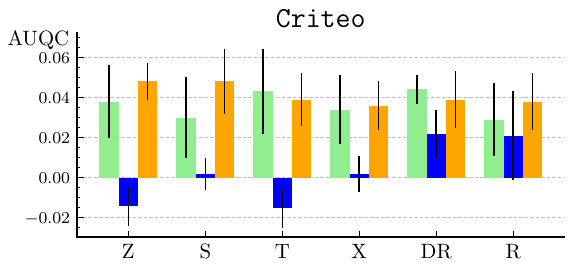}
\end{subfigure}
\hspace{5pt}

\vspace{10pt}

\hspace{5pt}
\begin{subfigure}[c]{0.33\textwidth}
    \centering
    \includegraphics[height=0.45\linewidth]{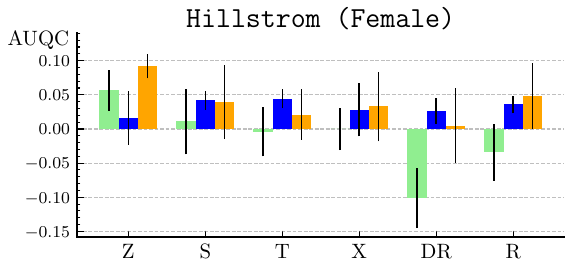}
\end{subfigure}
\hspace{5pt}
\begin{subfigure}[c]{0.33\textwidth}
    \centering
    \includegraphics[height=0.45\linewidth]{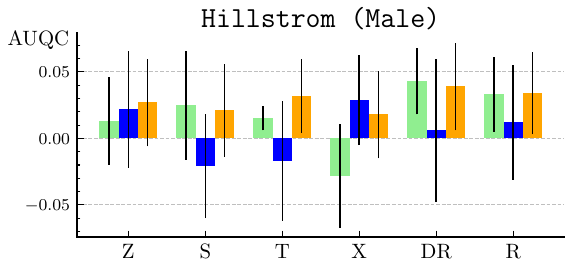}
\end{subfigure}%
\hspace{25pt}
\begin{subfigure}[c]{0.1\textwidth}
    \centering
    \includegraphics[width=\linewidth]{auqc_legend.pdf}
\end{subfigure}
\hspace{5pt}

% \vspace{-5pt}
\caption{\textit{Ranking Quality for Different Objectives and Metalearners.} For each metalearner, we compare three different objectives: point-, pair-, and listwise. We show performance in terms of AUQC $\pm$ one standard error, for five different data sets. As opposed to the figure in the main body, we do not scale the results here. Due to confidentiality reasons, we cannot share the raw results for the \texttt{Promotion} data set.
}
\label{fig:results_overview_overall_app}
\end{figure*}

We display the results presented in the main body of the text without normalizing the best value to $1$ in \cref{fig:results_overview_overall_app} and provide them in table format in \cref{tab:results_overview_overall}. 

\begin{table*}[]
    \centering
    \begin{tabular}{C{16pt}C{16pt}C{16pt}C{16pt}|R{60pt}R{60pt}R{60pt}R{60pt}R{60pt}}
    \toprule
        \multirow{2}{*}{\textbf{Meta}}  & \multicolumn{3}{c}{\textbf{Objective}} & \multicolumn{5}{c}{\textbf{Data}} \\
    \cmidrule(lr){2-4} \cmidrule(l){5-9}
        & \textit{Point} & \textit{Pair} & \multicolumn{1}{c}{\textit{List}} & \multicolumn{1}{r}{\texttt{Synthetic}} & \multicolumn{1}{r}{\texttt{Criteo}} & \multicolumn{1}{r}{\texttt{Hillstrom|\venus}} & \multicolumn{1}{r}{\texttt{Hillstrom|\mars}}  & \multicolumn{1}{r}{\texttt{Promotion}*} \\
    \midrule
        \multirow{3}{*}{\it Z}  & \cmark & \xmark & \xmark & \plus  0.183 {\scriptsize (0.027)}             & 	\plus  0.038 {\scriptsize (0.018)} & 	\plus  0.057 {\scriptsize (0.030)} & 	\plus  0.013 {\scriptsize (0.033)} & 	\plus  0.407 {\scriptsize (0.111)} 	 \\
                                & \xmark & \cmark & \xmark & \plus  0.137 {\scriptsize (0.016)}             & 	\minus 0.014 {\scriptsize (0.010)} & 	\plus  0.016 {\scriptsize (0.039)} & 	\plus  0.022 {\scriptsize (0.044)} & 	\plus  0.370 {\scriptsize (0.296)} 	 \\
                                & \xmark & \xmark & \cmark & \plus  {\bftab 0.242} {\scriptsize (0.026)}    & 	\plus  {\bftab 0.048} {\scriptsize (0.009)} & 	\plus  {\bftab 0.092} {\scriptsize (0.018)} & 	\plus  {\bftab 0.027} {\scriptsize (0.033)} & 	\plus  {\bftab 0.593} {\scriptsize (0.037)} 	 \\
    \midrule
        \multirow{3}{*}{\it S}  & \cmark & \xmark & \xmark & \plus  0.051 {\scriptsize (0.050)}             & 	\plus  0.030 {\scriptsize (0.020)} & 	\plus  0.011 {\scriptsize (0.048)} & 	\plus  {\bftab 0.025} {\scriptsize (0.041)} & 	\plus  0.370 {\scriptsize (0.074)} 	 \\
                                & \xmark & \cmark & \xmark & \plus  {\bftab 0.052} {\scriptsize (0.039)}    & 	\plus  0.002 {\scriptsize (0.008)} & 	\plus  {\bftab 0.042} {\scriptsize (0.014)} & 	\minus 0.021 {\scriptsize (0.039)} & 	\plus  {\bftab 0.593} {\scriptsize (0.296)} 	 \\
                                & \xmark & \xmark & \cmark & \plus  0.037 {\scriptsize (0.027)}             & 	\plus  {\bftab 0.048} {\scriptsize (0.016)} & 	\plus  0.039 {\scriptsize (0.054)} & 	\plus  0.021 {\scriptsize (0.035)} & 	\plus  0.222 {\scriptsize (0.148)} 	 \\
    \midrule
        \multirow{3}{*}{\it T}  & \cmark & \xmark & \xmark & \plus  0.054 {\scriptsize (0.049)}             & 	\plus  {\bftab 0.043} {\scriptsize (0.021)} & 	\minus 0.004 {\scriptsize (0.036)} & 	\plus  0.015 {\scriptsize (0.009)} & 	\plus  0.333 {\scriptsize (0.148)} 	 \\
                                & \xmark & \cmark & \xmark & \plus  0.046 {\scriptsize (0.044)}             & 	\minus 0.015 {\scriptsize (0.010)} & 	\plus  {\bftab 0.044} {\scriptsize (0.014)} & 	\minus 0.017 {\scriptsize (0.045)} & 	\plus  {\bftab 0.630} {\scriptsize (0.185)} 	 \\
                                & \xmark & \xmark & \cmark & \plus  {\bftab 0.058} {\scriptsize (0.043)}    & 	\plus  0.039 {\scriptsize (0.013)} & 	\plus  0.021 {\scriptsize (0.037)} & 	\plus  {\bftab 0.032} {\scriptsize (0.028)} & 	\plus  0.556 {\scriptsize (0.222)} 	 \\
    \midrule
        \multirow{3}{*}{\it X}  & \cmark & \xmark & \xmark & \plus  0.052 {\scriptsize (0.049)}             & 	\plus  0.034 {\scriptsize (0.017)} & 	\plus  -0.000 {\scriptsize (0.031)} & 	\minus 0.028 {\scriptsize (0.039)} & 	\plus  {\bftab 1.000} {\scriptsize (0.296)} 	 \\
                                & \xmark & \cmark & \xmark & \plus  0.054 {\scriptsize (0.042)}             & 	\plus  0.002 {\scriptsize (0.009)} & 	\plus  0.028 {\scriptsize (0.039)} & 	\plus  {\bftab 0.029} {\scriptsize (0.034)} & 	\plus  0.704 {\scriptsize (0.259)} 	 \\
                                & \xmark & \xmark & \cmark & \plus  {\bftab 0.061} {\scriptsize (0.038)}    & 	\plus  {\bftab 0.036} {\scriptsize (0.012)} & 	\plus  {\bftab 0.033} {\scriptsize (0.051)} & 	\plus  0.018 {\scriptsize (0.033)} & 	\plus  0.852 {\scriptsize (0.111)} 	 \\
    \midrule
        \multirow{3}{*}{\it DR} & \cmark & \xmark & \xmark & \plus  0.055 {\scriptsize (0.046)}             & 	\plus  {\bftab 0.044} {\scriptsize (0.007)} & 	\minus 0.101 {\scriptsize (0.044)} & 	\plus  {\bftab 0.043} {\scriptsize (0.025)} & 	\plus  {\bftab 0.667} {\scriptsize (0.111)} 	 \\
                                & \xmark & \cmark & \xmark & \plus  0.048 {\scriptsize (0.047)}             & 	\plus  0.022 {\scriptsize (0.012)} & 	\plus  {\bftab 0.027} {\scriptsize (0.019)} & 	\plus  0.006 {\scriptsize (0.054)} & 	\minus 0.481 {\scriptsize (0.222)} 	 \\
                                & \xmark & \xmark & \cmark & \plus  {\bftab 0.056} {\scriptsize (0.038)}    & 	\plus  0.039 {\scriptsize (0.014)} & 	\plus  0.005 {\scriptsize (0.055)} & 	\plus  0.039 {\scriptsize (0.033)} & 	\minus 0.148 {\scriptsize (0.222)} 	 \\
    \midrule
        \multirow{3}{*}{\it R}  & \cmark & \xmark & \xmark & \plus  0.054 {\scriptsize (0.046)}             & 	\plus  0.029 {\scriptsize (0.018)} & 	\minus 0.034 {\scriptsize (0.042)} & 	\plus  0.033 {\scriptsize (0.028)} & 	\plus  {\bftab 0.593} {\scriptsize (0.222)} 	 \\
                                & \xmark & \cmark & \xmark & \plus  0.050 {\scriptsize (0.047)}             & 	\plus  0.021 {\scriptsize (0.022)} & 	\plus  0.036 {\scriptsize (0.013)} & 	\plus  0.012 {\scriptsize (0.043)} & 	\minus 0.296 {\scriptsize (0.111)} 	 \\
                                & \xmark & \xmark & \cmark & \plus  {\bftab 0.056} {\scriptsize (0.039)}    & 	\plus  {\bftab 0.038} {\scriptsize (0.014)} & 	\plus  {\bftab 0.048} {\scriptsize (0.048)} & 	\plus  {\bftab 0.034} {\scriptsize (0.031)} & 	\minus 0.259 {\scriptsize (0.259)} 	 \\
    \bottomrule
    \multicolumn{9}{l}{\small \text{*For the \texttt{Promotion} data, we only present scaled results such that the best $\text{AUQC}=1$ due to reasons of confidentiality.}}
    \end{tabular}
    % \vspace{2pt}
    \caption{\textit{Ranking Quality for Different Objectives and Metalearners.} For each metalearner, we compare three different objectives: point-, pair-, and listwise. We show performance in terms of AUQC (with standard error in brackets), for the \texttt{Synthetic}, \texttt{Criteo}, and \texttt{Hillstrom} (\texttt{Women} (\venus) and \texttt{Men} (\mars) e-mail), and \texttt{Promotion} data. }
    \label{tab:results_overview_overall}
\end{table*}

\subsection{Analyzing Alternative Metrics (RQ2) and Design Choices (RQ3): Additional Results}
\label{ssec:rq2_3_additional}

First, we present additional results to support our investigation surrounding RQ2. For each metalearner, we visualize the trade-off between different metrics: the MSE (i.e., pointwise accuracy), Kendall $\tau$ (i.e., pairwise rank correlation), and AUQC (i.e., listwise ranking quality). This shows that models that perform well in terms of AUQC, also perform well in terms of Kendall $\tau$. Conversely, performance in terms of MSE does not seem related to AUQC or Kendall $\tau$. Finally, we also display these results in table format in \cref{tab:results_overview_synthetic}. 

\begin{figure*}[]
\centering
    \hspace{40pt}
    \begin{subfigure}[b]{0.22\textwidth}
        \centering
        \includegraphics[width=\textwidth]{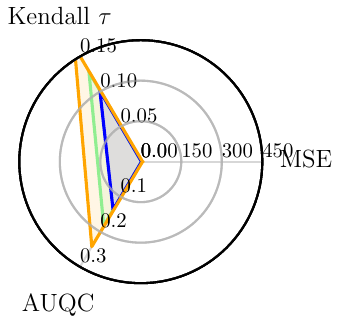}
        \caption{{Z-Learner}}
    \end{subfigure}
    \hfill
    \begin{subfigure}[b]{0.22\textwidth}
        \centering
        \includegraphics[width=\textwidth]{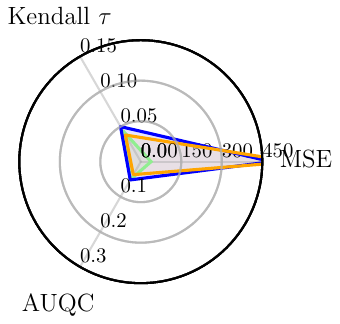}
        \caption{{S-Learner}}
    \end{subfigure}
    \hfill
    \begin{subfigure}[b]{0.22\textwidth}
        \centering
        \includegraphics[width=\textwidth]{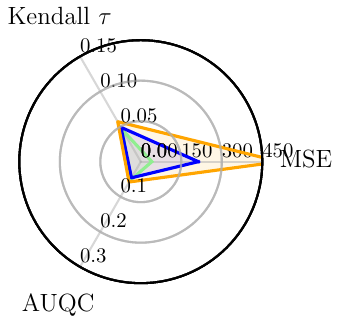}
        \caption{{T-Learner}}
    \end{subfigure}
    \hspace{40pt}

    \vspace{8pt}

    \hspace{40pt}
    \begin{subfigure}[b]{0.22\textwidth}
        \centering
        \includegraphics[width=\textwidth]{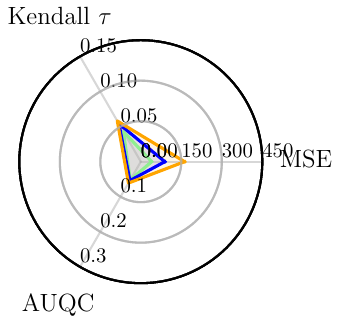}
        \caption{{X-Learner}}
    \end{subfigure}
    \hfill
    \begin{subfigure}[b]{0.22\textwidth}
        \centering
        \includegraphics[width=\textwidth]{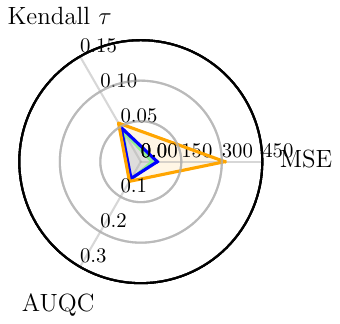}
        \caption{{DR-Learner}}
    \end{subfigure}
    \hfill
    \begin{subfigure}[b]{0.22\textwidth}
        \centering
        \includegraphics[width=\textwidth]{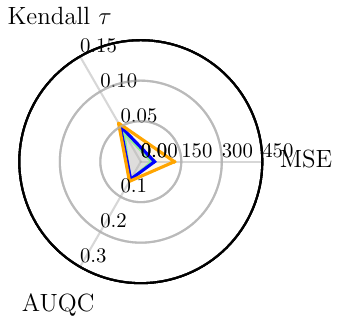}
        \caption{{R-Learner}}
    \end{subfigure}
    \hspace{40pt}

\vspace{-5pt}
\caption{\textit{Analyzing Performance Trade-offs on Synthetic Data.} For each metalearner, we compare three different objectives: \textcolor{green}{pointwise}, \textcolor{blue}{pairwise}, and \textcolor{orange}{listwise}. Using the \texttt{Synthetic} data set, we compare performance in terms of MSE (measuring pointwise accuracy), Kendall $\tau$ (measuring pairwise rank correlation), and AUQC (measuring global, listwise decision quality).}
\label{fig:results_overview_synthetic_radar}
\end{figure*}

\begin{table*}[]
    \centering
    \begin{tabular}{C{16pt}C{16pt}C{16pt}C{16pt}|R{75pt}R{75pt}R{75pt}}
    \toprule
        \multirow{2}{*}{\textbf{Meta}}  & \multicolumn{3}{c}{\textbf{Objective}} & \multicolumn{3}{c}{\textbf{Metric}} \\
    \cmidrule(lr){2-4} \cmidrule(l){5-7}
        & \textit{Point} & \textit{Pair} & \multicolumn{1}{c}{\textit{List}} & \multicolumn{1}{r}{MSE} & \multicolumn{1}{r}{Kendall $\tau$} & \multicolumn{1}{r}{AUQC} \\
    % \specialrule{\lightrulewidth}{2pt}{0.75pt}
    % \specialrule{\lightrulewidth}{0.75pt}{3pt}
    \midrule
        \multirow{3}{*}{\it Z}  & \cmark & \xmark & \xmark & \plus  {\bftab 2.474} {\scriptsize (0.255)}     & 	\plus  0.128 {\scriptsize (0.021)} & 	\plus  0.183 {\scriptsize (0.027)} \\
                                & \xmark & \cmark & \xmark & \plus  2.867 {\scriptsize (0.263)}     & 	\plus  0.101 {\scriptsize (0.013)} & 	\plus  0.137 {\scriptsize (0.016)} \\
                                & \xmark & \xmark & \cmark & \plus  6.465 {\scriptsize (1.260)}     & 	\plus  {\bftab 0.164} {\scriptsize (0.020)} & 	\plus  {\bftab 0.242} {\scriptsize (0.026)} \\

    \midrule
        \multirow{3}{*}{\it S}  & \cmark & \xmark & \xmark & \plus  {\bftab 37.916} {\scriptsize (3.557)}    & 	\plus  0.044 {\scriptsize (0.033)} & 	\plus  0.051 {\scriptsize (0.050)} \\
                                & \xmark & \cmark & \xmark & \plus  483.712 {\scriptsize (159.054)} & 	\plus  {\bftab 0.050} {\scriptsize (0.028)} & 	\plus  {\bftab 0.052} {\scriptsize (0.039)} \\
                                & \xmark & \xmark & \cmark & \plus  562.509 {\scriptsize (300.946)} & 	\plus  0.038 {\scriptsize (0.022)} & 	\plus  0.037 {\scriptsize (0.027)} \\

    \midrule
        \multirow{3}{*}{\it T}  & \cmark & \xmark & \xmark & \plus  {\bftab 41.014} {\scriptsize (3.852)}    & 	\plus  0.045 {\scriptsize (0.033)} & 	\plus  0.054 {\scriptsize (0.049)} \\
                                & \xmark & \cmark & \xmark & \plus  213.265 {\scriptsize (30.716)}  & 	\plus  0.049 {\scriptsize (0.030)} & 	\plus  0.046 {\scriptsize (0.044)} \\
                                & \xmark & \xmark & \cmark & \plus  513.908 {\scriptsize (94.592)}  & 	\plus  {\bftab 0.057} {\scriptsize (0.030)} & 	\plus  {\bftab 0.058} {\scriptsize (0.043)} \\

    \midrule
        \multirow{3}{*}{\it X}  & \cmark & \xmark & \xmark & \plus  {\bftab 40.417} {\scriptsize (3.697)}    & 	\plus  0.043 {\scriptsize (0.033)} & 	\plus  0.052 {\scriptsize (0.049)} \\
                                & \xmark & \cmark & \xmark & \plus  88.987 {\scriptsize (16.651)}   & 	\plus  0.054 {\scriptsize (0.030)} & 	\plus  0.054 {\scriptsize (0.042)} \\
                                & \xmark & \xmark & \cmark & \plus  162.146 {\scriptsize (40.750)}  & 	\plus  {\bftab 0.058} {\scriptsize (0.028)} & 	\plus  {\bftab 0.061} {\scriptsize (0.038)} \\

    \midrule
        \multirow{3}{*}{\it DR} & \cmark & \xmark & \xmark & \plus  {\bftab 42.007} {\scriptsize (4.017)}    & 	\plus  0.043 {\scriptsize (0.031)} & 	\plus  0.055 {\scriptsize (0.046)} \\
                                & \xmark & \cmark & \xmark & \plus  61.014 {\scriptsize (10.268)}   & 	\plus  0.049 {\scriptsize (0.032)} & 	\plus  0.048 {\scriptsize (0.047)} \\
                                & \xmark & \xmark & \cmark & \plus  311.236 {\scriptsize (105.607)} & 	\plus  {\bftab 0.055} {\scriptsize (0.028)} & 	\plus  {\bftab 0.056} {\scriptsize (0.038)} \\

    \midrule
        \multirow{3}{*}{\it R}  & \cmark & \xmark & \xmark & \plus  {\bftab 41.263} {\scriptsize (4.760)}    & 	\plus  0.042 {\scriptsize (0.031)} & 	\plus  0.054 {\scriptsize (0.046)} \\
                                & \xmark & \cmark & \xmark & \plus  51.008 {\scriptsize (12.094)}   & 	\plus  0.051 {\scriptsize (0.032)} & 	\plus  0.050 {\scriptsize (0.047)} \\
                                & \xmark & \xmark & \cmark & \plus  124.400 {\scriptsize (18.588)}  & 	\plus  {\bftab 0.055} {\scriptsize (0.028)} & 	\plus  {\bftab 0.056} {\scriptsize (0.039)} \\
    \bottomrule
    \end{tabular}
    \vspace{5pt}
    \caption{\textit{Analyzing Performance Trade-offs on Synthetic Data.} For each metalearner, we compare three different objectives: pointwise, pairwise, and listwise. Using the \texttt{Synthetic} data set, we compare performance in terms of MSE (measuring pointwise accuracy), Kendall $\tau$ (measuring pairwise rank correlation), and AUQC (measuring global, listwise decision quality). For each, we show the standard error in brackets.}
    \label{tab:results_overview_synthetic}
\end{table*}

\subsection{Sensitivity Analysis}
\label{ssec:sensitivity_analysis}

To single out the effect of the analyzed design choice, we train with default hyperparameters for each training objective and metalearner. We explore three hyperparameters: (1) the number of sampling iterations $k$, i.e., the number of sampled pairs per instance, (2) the sigmoid parameter $\sigma$ controlling the steepness of the comparison in the construction of the pairwise score (see \cref{eq:pairwise_score}), and (3) whether we normalize the ranking score by feeding it to a logistic sigmoid and constraining it to $[0, 1]$. Related work has shown that this normalization might effectively serve as a form of regularization and help with overfitting \citep{du2019improve}.

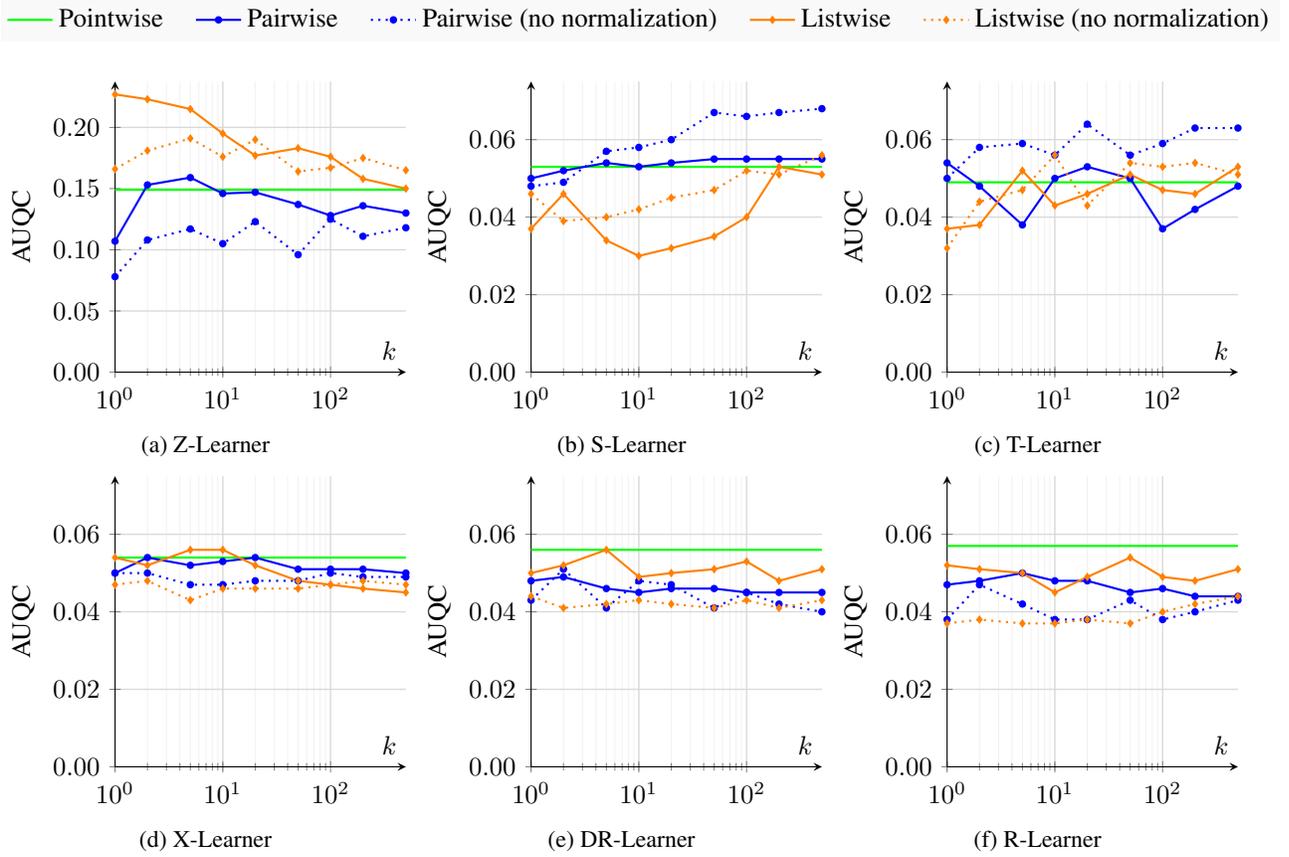
\begin{figure*}
    \centering
    \begin{subfigure}[b]{\linewidth}
    \centering
        \begin{tikzpicture}
        \begin{axis}[
          hide axis, 
          axis lines=middle,
          grid=both,
          grid style={line width=.1pt, draw=gray!10},
          major grid style={line width=0.5pt,draw=gray!30},
          xmin=1,
          xmax=1000,
          xmode=log,
          ymin=0.075,
          ymax=0.325,
          xlabel=$k$,
          % ylabel=AUQC,
          x label style={at={(1, 0.01)}},
          ylabel style={rotate=-90},
          ylabel near ticks, 
          xtick={1, 10, 100, 1000},
          ytick={0.0,0.1,...,0.4},
          hide obscured y ticks=false,
          width=\linewidth,
          height=\linewidth,
          legend style={
              draw=none, fill=gray!5, inner sep=2pt, font={},
              /tikz/every even column/.append style={column sep=10pt},
          },
          legend cell align={left},
          legend columns=5,
          every axis plot/.append style={thick},
        ]
        % \footnotesize
        
        \addlegendimage{color=green, mark=none}
        \addlegendentry{Pointwise}
        
        \addlegendimage{color=blue, mark=*, mark options={scale=0.5, solid}}
        \addlegendentry{Pairwise}
        
        \addlegendimage{color=blue, mark=*, dotted, mark options={scale=0.5, solid}}
        % \addlegendentry{Pairwise' No Norm}
        \addlegendentry{Pairwise (\text{no normalization})}
        
        \addlegendimage{color=orange, mark=diamond*, mark options={scale=0.5, solid}}
        \addlegendentry{Listwise}
        
        \addlegendimage{color=orange, mark=diamond*, dotted, mark options={scale=0.5, solid}}
        % \addlegendentry{Listwise' No Norm}
        \addlegendentry{Listwise (\text{no normalization})}
        \end{axis}
        \end{tikzpicture}
    \end{subfigure}
    \vspace{3pt}
    
    \begin{subfigure}[b]{0.33\linewidth}
        \centering
        % \resizebox{\linewidth}{!}{%
        \begin{tikzpicture}
        \begin{axis}[
          axis lines=middle,
          grid=both,
          grid style={line width=.1pt, draw=gray!10},
          major grid style={line width=0.5pt,draw=gray!30},
          xmin=1,
          xmax=500,
          xmode=log,
          ymin=0.0,
          ymax=0.2375,
          xlabel=$k$,
          ylabel=AUQC,
          x label style={at={(1, 0.01)}},
          ylabel style={rotate=-90},
          ylabel near ticks, 
          xtick={1, 10, 100},
          ytick={0.0, 0.05, ..., 0.4},
          hide obscured x ticks=false,
          hide obscured y ticks=false,
          width=\linewidth,
          height=\linewidth,
          legend style={
              draw=none, fill=gray!5, inner sep=2pt,
          },
          legend cell align={left},
          every axis plot/.append style={thick},
          y tick label style={
            /pgf/number format/.cd,
            fixed,
            fixed zerofill,
            precision=2,
            /tikz/.cd
          }
        ]
        % \footnotesize
        \addplot[color=green, mark=none, error bars/.cd, y dir=both, y explicit]
            coordinates {
            (1, 0.149 )
            (1000, 0.149 )
            };
        \addplot[color=blue, mark=*, mark options={scale=0.5, solid}]
            coordinates {
            (1, 0.107 )
            (2, 0.153 )
            (5, 0.159 )
            (10, 0.146 )
            (20, 0.147 )
            (50, 0.137 )
            (100, 0.128 )
            (200, 0.136 )
            (500, 0.130 )
            % (1000, 0.133 )
            };
        % \addlegendentry{Pairwise}
        \addplot[color=blue, mark=*, dotted, mark options={scale=0.5, solid}]
            coordinates {
            (1, 0.078 )
            (2, 0.108 )
            (5, 0.117 )
            (10, 0.105 )
            (20, 0.123 )
            (50, 0.096 )
            (100, 0.125 )
            (200, 0.111 )
            (500, 0.118 )
            % (1000, )
            };
        % \addlegendentry{Pairwise'}
        \addplot[color=orange, mark=diamond*, mark options={scale=0.5, solid}]
            coordinates {
            (1, 0.227 )
            (2, 0.223 )
            (5, 0.215 )
            (10, 0.195 )
            (20, 0.177 )
            (50, 0.183 )
            (100, 0.176 )
            (200, 0.158 )
            (500, 0.150 )
            % (1000, 0.176 )
            };
        % \addlegendentry{Listwise}
        \addplot[color=orange, mark=diamond*, dotted, mark options={scale=0.5, solid}]
            coordinates {
            (1, 0.166 )
            (2, 0.181 )
            (5, 0.191 )
            (10, 0.176 )
            (20, 0.190 )
            (50, 0.164 )
            (100, 0.167 )
            (200, 0.175 )
            (500, 0.165 )
            % (1000, )
            };
        % \addlegendentry{Listwise'}
        \end{axis}
        \end{tikzpicture}
        % }
        \caption{Z-Learner}
    \end{subfigure}
    \hfill
    \begin{subfigure}[b]{0.33\textwidth}
        \centering
        \begin{tikzpicture}
        \begin{axis}[
          axis lines=middle,
          grid=both,
          grid style={line width=.1pt, draw=gray!10},
          major grid style={line width=0.5pt,draw=gray!30},
          xmin=1,
          xmax=500,
          xmode=log,
          ymin=0.0,
          ymax=0.075,
          xlabel=$k$,
          ylabel=AUQC,
          x label style={at={(1, 0.01)}},
          ylabel style={rotate=-90},
          ylabel near ticks, 
          xtick={1, 10, 100},
          ytick={0.0, 0.02, ..., 0.4},
          hide obscured x ticks=false,
          hide obscured y ticks=false,
          width=\linewidth,
          height=\linewidth,
          legend style={
              draw=none, fill=gray!5, inner sep=2pt,
          },
          legend cell align={left},
          every axis plot/.append style={thick},
          y tick label style={
            /pgf/number format/.cd,
            fixed,
            fixed zerofill,
            precision=2,
            /tikz/.cd
          },
        ]
        % \footnotesize
        \addplot[color=green, mark=none, ]
            coordinates {
            (1, 0.053 )
            (1000, 0.053 )
            };
        % \addlegendentry{Pointwise}
        \addplot[color=blue, mark=*, mark options={scale=0.5, solid}]
            coordinates {
            (1, 0.050 )
            (2, 0.052 )
            (5, 0.054 )
            (10, 0.053 )
            (20, 0.054 )
            (50, 0.055 )
            (100, 0.055 )
            (200, 0.055 )
            (500, 0.055 )
            % (1000, 0.057 )
            };
        % \addlegendentry{Pairwise}
        \addplot[color=blue, mark=*, dotted, mark options={scale=0.5, solid}]
            coordinates {
            (1, 0.048 )
            (2, 0.049 )
            (5, 0.057 )
            (10, 0.058 )
            (20, 0.060 )
            (50, 0.067 )
            (100, 0.066 )
            (200, 0.067 )
            (500, 0.068 )
            % (1000, )
            };
        % \addlegendentry{Pairwise'}
        \addplot[color=orange, mark=diamond*, mark options={scale=0.5, solid}]
            coordinates {
            (1, 0.037 )
            (2, 0.046 )
            (5, 0.034 )
            (10, 0.030 )
            (20, 0.032 )
            (50, 0.035 )
            (100, 0.040 )
            (200, 0.053 )
            (500, 0.051 )
            % (1000, 0.032 )
            };
        % \addlegendentry{Listwise}
        \addplot[color=orange, mark=diamond*, dotted, mark options={scale=0.5, solid}]
            coordinates {
            (1, 0.046 )
            (2, 0.039 )
            (5, 0.040 )
            (10, 0.042 )
            (20, 0.045 )
            (50, 0.047 )
            (100, 0.052 )
            (200, 0.051 )
            (500, 0.056 )
            % (1000, )
            };
        % \addlegendentry{Listwise'}
        \end{axis}
        \end{tikzpicture}
        \caption{S-Learner}
    \end{subfigure}
    \hfill
    \begin{subfigure}[b]{0.33\textwidth}
    \centering
        \begin{tikzpicture}
        \begin{axis}[
          axis lines=middle,
          grid=both,
          grid style={line width=.1pt, draw=gray!10},
          major grid style={line width=0.5pt,draw=gray!30},
          xmin=1,
          xmax=500,
          xmode=log,
          ymin=0.0,
          ymax=0.075,
          xlabel=$k$,
          ylabel=AUQC,
          x label style={at={(1, 0.01)}},
          ylabel style={rotate=-90},
          ylabel near ticks, 
          xtick={1, 10, 100},
          ytick={0.0, 0.02, ..., 0.4},
          hide obscured x ticks=false,
          hide obscured y ticks=false,
          width=\linewidth,
          height=\linewidth,
          legend style={
              draw=none, fill=gray!5, inner sep=2pt,
          },
          legend cell align={left},
          every axis plot/.append style={thick},
          y tick label style={
            /pgf/number format/.cd,
            fixed,
            fixed zerofill,
            precision=2,
            /tikz/.cd
          },
        ]
        % \footnotesize
        \addplot[color=green, mark=none, ]
            coordinates {
            (1, 0.049 )
            (1000, 0.049 )
            };
        % \addlegendentry{Pointwise}
        \addplot[color=blue, mark=*, mark options={scale=0.5, solid}]
            coordinates {
            (1, 0.054 )
            (2, 0.048 )
            (5, 0.038 )
            (10, 0.050 )
            (20, 0.053 )
            (50, 0.050 )
            (100, 0.037 )
            (200, 0.042 )
            (500, 0.048 )
            % (1000, 0.043 )
            };
        % \addlegendentry{Pairwise}
        \addplot[color=blue, mark=*, dotted, mark options={scale=0.5, solid}]
            coordinates {
            (1, 0.050 )
            (2, 0.058 )
            (5, 0.059 )
            (10, 0.056 )
            (20, 0.064 )
            (50, 0.056 )
            (100, 0.059 )
            (200, 0.063 )
            (500, 0.063 )
            % (1000, )
            };
        % \addlegendentry{Pairwise'}
        \addplot[color=orange, mark=diamond*, mark options={scale=0.5, solid}]
            coordinates {
            (1, 0.037 )
            (2, 0.038 )
            (5, 0.052 )
            (10, 0.043 )
            (20, 0.046 )
            (50, 0.051 )
            (100, 0.047 )
            (200, 0.046 )
            (500, 0.053 )
            % (1000, 0.055 )
            };
        % \addlegendentry{Listwise}
        \addplot[color=orange, mark=diamond*, dotted, mark options={scale=0.5, solid}]
            coordinates {
            (1, 0.032 )
            (2, 0.044 )
            (5, 0.047 )
            (10, 0.056 )
            (20, 0.043 )
            (50, 0.054 )
            (100, 0.053 )
            (200, 0.054 )
            (500, 0.051 )
            % (1000, )
            };
        % \addlegendentry{Listwise'}
        \end{axis}
        \end{tikzpicture}
        \caption{T-Learner}
    \end{subfigure}
    
    \vspace{5pt}
    
    \begin{subfigure}[b]{0.33\linewidth}
        \centering
        % \resizebox{\linewidth}{!}{%
        \begin{tikzpicture}
        \begin{axis}[
          axis lines=middle,
          grid=both,
          grid style={line width=.1pt, draw=gray!10},
          major grid style={line width=0.5pt,draw=gray!30},
          xmin=1,
          xmax=500,
          xmode=log,
          ymin=0.0,
          ymax=0.075,
          xlabel=$k$,
          ylabel=AUQC,
          x label style={at={(1, 0.01)}},
          ylabel style={rotate=-90},
          ylabel near ticks, 
          xtick={1, 10, 100},
          ytick={0.0, 0.02, ..., 0.4},
          hide obscured x ticks=false,
          hide obscured y ticks=false,
          width=\linewidth,
          height=\linewidth,
          legend style={
              draw=none, fill=gray!5, inner sep=2pt,
          },
          legend cell align={left},
          every axis plot/.append style={thick},
          y tick label style={
            /pgf/number format/.cd,
            fixed,
            fixed zerofill,
            precision=2,
            /tikz/.cd
          },
        ]
        % \footnotesize
        \addplot[color=green, mark=none, ]
            coordinates {
            (1, 0.054 )
            (1000, 0.054 )
            };
        % \addlegendentry{Pointwise}
        \addplot[color=blue, mark=*, mark options={scale=0.5, solid}]
            coordinates {
            (1, 0.050 )
            (2, 0.054 )
            (5, 0.052 )
            (10, 0.053 )
            (20, 0.054 )
            (50, 0.051 )
            (100, 0.051 )
            (200, 0.051 )
            (500, 0.050 )
            % (1000, 0.051 )
            };
        % \addlegendentry{Pairwise}
        \addplot[color=blue, mark=*, dotted, mark options={scale=0.5, solid}]
            coordinates {
            (1, 0.050 )
            (2, 0.050 )
            (5, 0.047 )
            (10, 0.047 )
            (20, 0.048 )
            (50, 0.048 )
            (100, 0.050 )
            (200, 0.049 )
            (500, 0.049 )
            % (1000, )
            };
        % \addlegendentry{Pairwise'}
        \addplot[color=orange, mark=diamond*, mark options={scale=0.5, solid}]
            coordinates {
            (1, 0.054 )
            (2, 0.052 )
            (5, 0.056 )
            (10, 0.056 )
            (20, 0.052 )
            (50, 0.048 )
            (100, 0.047 )
            (200, 0.046 )
            (500, 0.045 )
            % (1000, 0.047 )
            };
        % \addlegendentry{Listwise}
        \addplot[color=orange, mark=diamond*, dotted, mark options={scale=0.5, solid}]
            coordinates {
            (1, 0.047 )
            (2, 0.048 )
            (5, 0.043 )
            (10, 0.046 )
            (20, 0.046 )
            (50, 0.046 )
            (100, 0.047 )
            (200, 0.048 )
            (500, 0.047 )
            % (1000, )
            };
        % \addlegendentry{Listwise'}
        \end{axis}
        \end{tikzpicture}
        % }
        \caption{X-Learner}
    \end{subfigure}
    \hfill
    \begin{subfigure}[b]{0.33\textwidth}
        \centering
        \begin{tikzpicture}
        \begin{axis}[
          axis lines=middle,
          grid=both,
          grid style={line width=.1pt, draw=gray!10},
          major grid style={line width=0.5pt,draw=gray!30},
          xmin=1,
          xmax=500,
          xmode=log,
          ymin=0.0,
          ymax=0.075,
          xlabel=$k$,
          ylabel=AUQC,
          x label style={at={(1, 0.01)}},
          ylabel style={rotate=-90},
          ylabel near ticks, 
          xtick={1, 10, 100},
          ytick={0.0, 0.02, ..., 0.4},
          hide obscured x ticks=false,
          hide obscured y ticks=false,
          width=\linewidth,
          height=\linewidth,
          legend style={
              draw=none, fill=gray!5, inner sep=2pt,
          },
          legend cell align={left},
          every axis plot/.append style={thick},
          y tick label style={
            /pgf/number format/.cd,
            fixed,
            fixed zerofill,
            precision=2,
            /tikz/.cd
          },
        ]
        % \footnotesize
        \addplot[color=green, mark=none, ]
            coordinates {
            (1, 0.056 )
            (1000, 0.056 )
            };
        % \addlegendentry{Pointwise}
        \addplot[color=blue, mark=*, mark options={scale=0.5, solid}]
            coordinates {
            (1, 0.048 )
            (2, 0.049 )
            (5, 0.046 )
            (10, 0.045 )
            (20, 0.046 )
            (50, 0.046 )
            (100, 0.045 )
            (200, 0.045 )
            (500, 0.045 )
            % (1000, 0.044 )
            };
        % \addlegendentry{Pairwise}
        \addplot[color=blue, mark=*, dotted, mark options={scale=0.5, solid}]
            coordinates {
            (1, 0.043 )
            (2, 0.051 )
            (5, 0.041 )
            (10, 0.048 )
            (20, 0.047 )
            (50, 0.041 )
            (100, 0.045 )
            (200, 0.042 )
            (500, 0.040 )
            % (1000, )
            };
        % \addlegendentry{Pairwise'}
        \addplot[color=orange, mark=diamond*, mark options={scale=0.5, solid}]
            coordinates {
            (1, 0.050 )
            (2, 0.052 )
            (5, 0.056 )
            (10, 0.049 )
            (20, 0.050 )
            (50, 0.051 )
            (100, 0.053 )
            (200, 0.048 )
            (500, 0.051 )
            % (1000, 0.055  )
            };
        % \addlegendentry{Listwise}
        \addplot[color=orange, mark=diamond*, dotted, mark options={scale=0.5, solid}]
            coordinates {
            (1, 0.044 )
            (2, 0.041 )
            (5, 0.042 )
            (10, 0.043 )
            (20, 0.042 )
            (50, 0.041 )
            (100, 0.043 )
            (200, 0.041 )
            (500, 0.043 )
            % (1000, )
            };
        % \addlegendentry{Listwise'}
        \end{axis}
        \end{tikzpicture}
        \caption{DR-Learner}
    \end{subfigure}
    \hfill
    \begin{subfigure}[b]{0.33\textwidth}
    \centering
        \begin{tikzpicture}
        \begin{axis}[
          axis lines=middle,
          grid=both,
          grid style={line width=.1pt, draw=gray!10},
          major grid style={line width=0.5pt,draw=gray!30},
          xmin=1,
          xmax=500,
          xmode=log,
          ymin=0.0,
          ymax=0.075,
          xlabel=$k$,
          ylabel=AUQC,
          x label style={at={(1, 0.01)}},
          ylabel style={rotate=-90},
          ylabel near ticks, 
          xtick={1, 10, 100},
          ytick={0.0, 0.02, ..., 0.4},
          hide obscured x ticks=false,
          hide obscured y ticks=false,
          width=\linewidth,
          height=\linewidth,
          legend style={
              draw=none, fill=gray!5, inner sep=2pt,
          },
          legend cell align={left},
          every axis plot/.append style={thick},
          y tick label style={
            /pgf/number format/.cd,
            fixed,
            fixed zerofill,
            precision=2,
            /tikz/.cd
          },
        ]
        % \footnotesize
        \addplot[color=green, mark=none, ]
            coordinates {
            (1, 0.057 )
            (1000, 0.057 )
            };
        % \addlegendentry{Pointwise}
        \addplot[color=blue, mark=*, mark options={scale=0.5, solid}]
            coordinates {
            (1, 0.047 )
            (2, 0.048 )
            (5, 0.050 )
            (10, 0.048 )
            (20, 0.048 )
            (50, 0.045 )
            (100, 0.046 )
            (200, 0.044 )
            (500, 0.044 )
            % (1000, 0.043 )
            };
        % \addlegendentry{Pairwise}
        \addplot[color=blue, mark=*, dotted, mark options={scale=0.5, solid}]
            coordinates {
            (1, 0.038 )
            (2, 0.047 )
            (5, 0.042 )
            (10, 0.038 )
            (20, 0.038 )
            (50, 0.043 )
            (100, 0.038 )
            (200, 0.040 )
            (500, 0.043 )
            % (1000, )
            };
        % \addlegendentry{Pairwise'}
        \addplot[color=orange, mark=diamond*, mark options={scale=0.5, solid}]
            coordinates {
            (1, 0.052 )
            (2, 0.051 )
            (5, 0.050 )
            (10, 0.045 )
            (20, 0.049 )
            (50, 0.054 )
            (100, 0.049 )
            (200, 0.048 )
            (500, 0.051 )
            % (1000, 0.044 )
            };
        % \addlegendentry{Listwise}
        \addplot[color=orange, mark=diamond*, dotted, mark options={scale=0.5, solid}]
            coordinates {
            (1, 0.037 )
            (2, 0.038 )
            (5, 0.037 )
            (10, 0.037 )
            (20, 0.038 )
            (50, 0.037 )
            (100, 0.040 )
            (200, 0.042 )
            (500, 0.044 )
            % (1000, )
            };
        % \addlegendentry{Listwise'}
        \end{axis}
        \end{tikzpicture}
        \caption{R-Learner}
    \end{subfigure}
\caption{\textit{What Is the Effect of the Number of Sampling Iterations $k$?} We show performance in terms of AUQC (higher is better) for the different metalearners on the \texttt{Synthetic} data set. We fix the sigmoid parameter $\sigma = 1$ and train with default hyperparameters.}
\label{fig:experiment_synthetic_sampling_iterations_app}
\end{figure*}

In \cref{fig:experiment_synthetic_sampling_iterations_app}, we vary the number of sampled pairs $k$ in the ranking objectives, for each ranking metalearner, and compare it to the pointwise model on the \texttt{Synthetic} data. We also show results for the metalearners with normalization of the score (i.e., constraining the model to outputs between zero and one) and without normalization. Somewhat surprisingly, with normalization, we see that increasing $k$ does not yield better results. Conversely, without normalization, increasing $k$ does in fact improve performance for most metalearners. Nevertheless, for the Z-, X-, DR-, and R-Learner, the best performance is achieved with normalization and $k=1$--the same settings used in the experiments in the main body (i.e., \cref{tab:results_overview_overall}). For the S- and T-Learner, deviating from these settings might improve performance. Overall, we see that for each metalearner and objective, we can obtain the same performance or better than the pointwise equivalent given that the right hyperparameters are chosen. This insight provides another validation of our proposed approach.

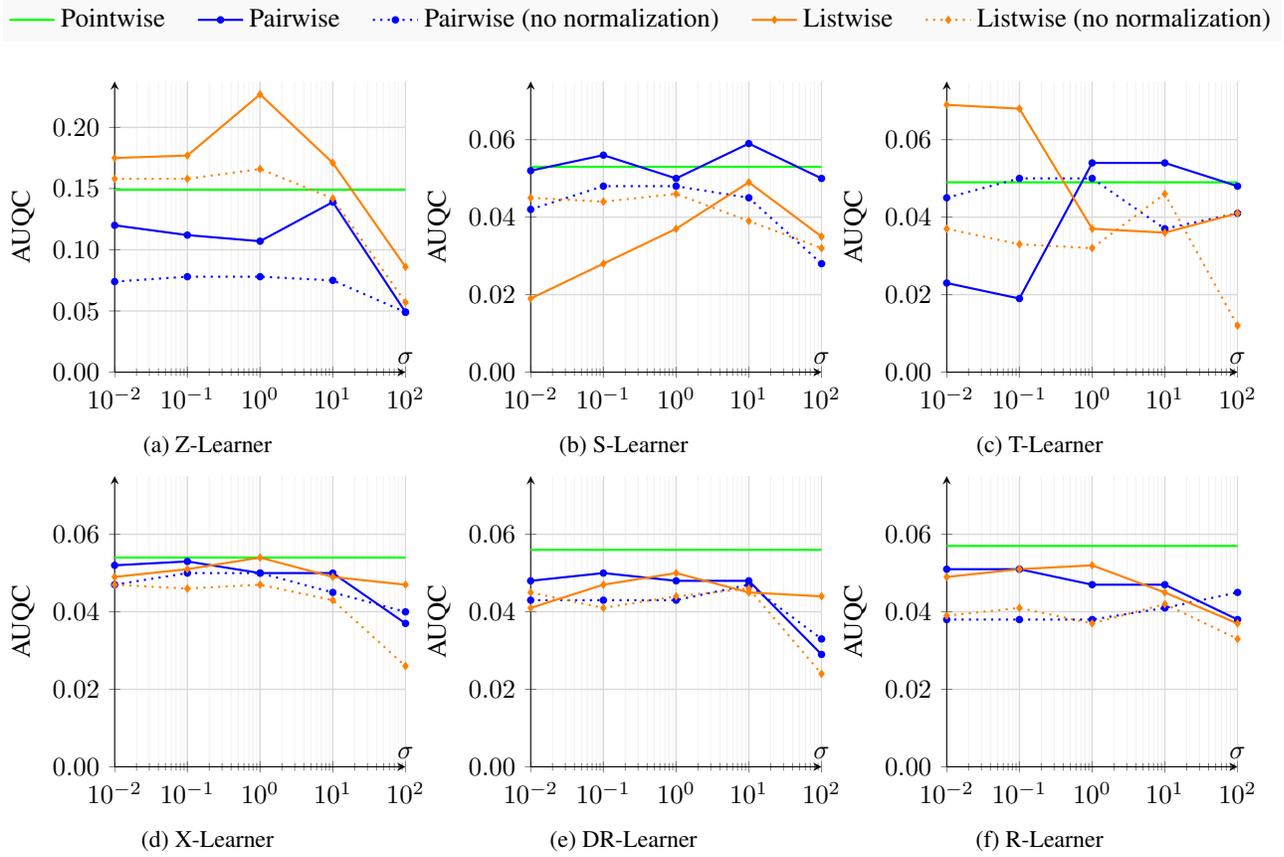
\begin{figure*}
    \centering
    \begin{subfigure}[b]{\linewidth}
    \centering
        \begin{tikzpicture}
        \begin{axis}[
          hide axis, 
          axis lines=left,
          grid=both,
          grid style={line width=.1pt, draw=gray!10},
          major grid style={line width=0.5pt,draw=gray!30},
          xmin=0.01,
          xmax=100,
          xmode=log,
          ymin=0.075,
          ymax=0.325,
          xlabel=$k$,
          % ylabel=AUQC,
          x label style={at={(1, 0.01)}},
          ylabel style={rotate=-90},
          ylabel near ticks, 
          xtick={1, 10, 100, 1000},
          ytick={0.0,0.1,...,0.4},
          hide obscured y ticks=false,
          width=\linewidth,
          height=\linewidth,
          legend style={
              draw=none, fill=gray!5, inner sep=2pt, font={},
              /tikz/every even column/.append style={column sep=10pt},
          },
          legend cell align={left},
          legend columns=5,
          every axis plot/.append style={thick},
        ]
        % \footnotesize
        
        \addlegendimage{color=green, mark=none}
        \addlegendentry{Pointwise}
        
        \addlegendimage{color=blue, mark=*, mark options={scale=0.5, solid}}
        \addlegendentry{Pairwise}
        
        \addlegendimage{color=blue, mark=*, dotted, mark options={scale=0.5, solid}}
        % \addlegendentry{Pairwise' No Norm}
        \addlegendentry{Pairwise (\text{no normalization})}
        
        \addlegendimage{color=orange, mark=diamond*, mark options={scale=0.5, solid}}
        \addlegendentry{Listwise}
        
        \addlegendimage{color=orange, mark=diamond*, dotted, mark options={scale=0.5, solid}}
        % \addlegendentry{Listwise' No Norm}
        \addlegendentry{Listwise (\text{no normalization})}
        \end{axis}
        \end{tikzpicture}
    \end{subfigure}
    \vspace{3pt}
    
    \begin{subfigure}[b]{0.33\linewidth}
        \centering
        % \resizebox{\linewidth}{!}{%
        \begin{tikzpicture}
        \begin{axis}[
          axis lines=left,
          grid=both,
          grid style={line width=.1pt, draw=gray!10},
          major grid style={line width=0.5pt,draw=gray!30},
          xmin=0.01,
          xmax=100,
          xmode=log,
          ymin=0.0,
          ymax=0.2375,
          xlabel=$\sigma$,
          ylabel=AUQC,
          x label style={at={(1, 0.10)}},
          ylabel style={rotate=-90},
          ylabel near ticks, 
          xtick={0.01, 0.1, 1, 10, 100},
          ytick={0.0, 0.05, ..., 0.4},
          hide obscured x ticks=false,
          hide obscured y ticks=false,
          width=\linewidth,
          height=\linewidth,
          legend style={
              draw=none, fill=gray!5, inner sep=2pt,
          },
          legend cell align={left},
          every axis plot/.append style={thick},
          y tick label style={
            /pgf/number format/.cd,
            fixed,
            fixed zerofill,
            precision=2,
            /tikz/.cd
          },
        ]
        % \footnotesize
        \addplot[color=green, mark=none, error bars/.cd, y dir=both, y explicit]
            coordinates {
            (0.01, 0.149 )
            (100, 0.149 )
            };
        \addplot[color=blue, mark=*, mark options={scale=0.5, solid}]
            coordinates {
            (0.01, 0.120 )
            (0.1, 0.112 )
            (1, 0.107 )
            (10, 0.139 )
            (100, 0.049 )
            };
        % \addlegendentry{Pairwise}
        \addplot[color=blue, mark=*, dotted, mark options={scale=0.5, solid}]
            coordinates {
            (0.01, 0.074 )
            (0.1, 0.078 )
            (1, 0.078 )
            (10, 0.075 )
            (100, 0.049 )
            };
        % \addlegendentry{Pairwise'}
        \addplot[color=orange, mark=diamond*, mark options={scale=0.5, solid}]
            coordinates {
            (0.01, 0.175 )
            (0.1, 0.177 )
            (1, 0.227 )
            (10, 0.171 )
            (100, 0.086 )
            };
        % \addlegendentry{Listwise}
        \addplot[color=orange, mark=diamond*, dotted, mark options={scale=0.5, solid}]
            coordinates {
            (0.01, 0.158 )
            (0.1, 0.158 )
            (1, 0.166 )
            (10, 0.142 )
            (100, 0.057 )
            };
        % \addlegendentry{Listwise'}
        \end{axis}
        \end{tikzpicture}
        % }
        \caption{Z-Learner}
    \end{subfigure}
    \hfill
    \begin{subfigure}[b]{0.33\textwidth}
        \centering
        \begin{tikzpicture}
        \begin{axis}[
          axis lines=left,
          grid=both,
          grid style={line width=.1pt, draw=gray!10},
          major grid style={line width=0.5pt,draw=gray!30},
          xmin=0.01,
          xmax=100,
          xmode=log,
          ymin=0.0,
          ymax=0.075,
          xlabel=$\sigma$,
          ylabel=AUQC,
          x label style={at={(1, 0.10)}},
          ylabel style={rotate=-90},
          ylabel near ticks, 
          xtick={0.01, 0.1, 1, 10, 100},
          ytick={0.0, 0.02, ..., 0.4},
          hide obscured x ticks=false,
          hide obscured y ticks=false,
          width=\linewidth,
          height=\linewidth,
          legend style={
              draw=none, fill=gray!5, inner sep=2pt,
          },
          legend cell align={left},
          every axis plot/.append style={thick},
          y tick label style={
            /pgf/number format/.cd,
            fixed,
            fixed zerofill,
            precision=2,
            /tikz/.cd
          },
        ]
        % \footnotesize
        \addplot[color=green, mark=none, ]
            coordinates {
            (0.01, 0.053 )
            (100, 0.053 )
            };
        % \addlegendentry{Pointwise}
        \addplot[color=blue, mark=*, mark options={scale=0.5, solid}]
            coordinates {
            (0.01, 0.052 )
            (0.1, 0.056 )
            (1, 0.050 )
            (10, 0.059 )
            (100, 0.050 )
            };
        % \addlegendentry{Pairwise}
        \addplot[color=blue, mark=*, dotted, mark options={scale=0.5, solid}]
            coordinates {
            (0.01, 0.042 )
            (0.1, 0.048 )
            (1, 0.048 )
            (10, 0.045 )
            (100, 0.028 )
            };
        % \addlegendentry{Pairwise'}
        \addplot[color=orange, mark=diamond*, mark options={scale=0.5, solid}]
            coordinates {
            (0.01, 0.019 )
            (0.1, 0.028 )
            (1, 0.037 )
            (10, 0.049 )
            (100, 0.035 )
            };
        % \addlegendentry{Listwise}
        \addplot[color=orange, mark=diamond*, dotted, mark options={scale=0.5, solid}]
            coordinates {
            (0.01, 0.045 )
            (0.1, 0.044 )
            (1, 0.046 )
            (10, 0.039 )
            (100, 0.032 )
            };
        % \addlegendentry{Listwise'}
        \end{axis}
        \end{tikzpicture}
        \caption{S-Learner}
    \end{subfigure}
    \hfill
    \begin{subfigure}[b]{0.33\textwidth}
    \centering
        \begin{tikzpicture}
        \begin{axis}[
          axis lines=left,
          grid=both,
          grid style={line width=.1pt, draw=gray!10},
          major grid style={line width=0.5pt,draw=gray!30},
          xmin=0.01,
          xmax=100,
          xmode=log,
          ymin=0.0,
          ymax=0.075,
          xlabel=$\sigma$,
          ylabel=AUQC,
          x label style={at={(1, 0.10)}},
          ylabel style={rotate=-90},
          ylabel near ticks, 
          xtick={0.01, 0.1, 1, 10, 100},
          ytick={0.0, 0.02, ..., 0.4},
          hide obscured x ticks=false,
          hide obscured y ticks=false,
          width=\linewidth,
          height=\linewidth,
          legend style={
              draw=none, fill=gray!5, inner sep=2pt,
          },
          legend cell align={left},
          every axis plot/.append style={thick},
          y tick label style={
            /pgf/number format/.cd,
            fixed,
            fixed zerofill,
            precision=2,
            /tikz/.cd
          },
        ]
        % \footnotesize
        \addplot[color=green, mark=none, ]
            coordinates {
            (0.01, 0.049 )
            (100, 0.049 )
            };
        % \addlegendentry{Pointwise}
        \addplot[color=blue, mark=*, mark options={scale=0.5, solid}]
            coordinates {
            (0.01, 0.023 )
            (0.1, 0.019 )
            (1, 0.054 )
            (10, 0.054 )
            (100, 0.048 )
            };
        % \addlegendentry{Pairwise}
        \addplot[color=blue, mark=*, dotted, mark options={scale=0.5, solid}]
            coordinates {
            (0.01, 0.045 )
            (0.1, 0.050 )
            (1, 0.050 )
            (10, 0.037 )
            (100, 0.041 )
            };
        % \addlegendentry{Pairwise'}
        \addplot[color=orange, mark=diamond*, mark options={scale=0.5, solid}]
            coordinates {
            (0.01, 0.069 )
            (0.1, 0.068 )
            (1, 0.037 )
            (10, 0.036 )
            (100, 0.041 )
            };
        % \addlegendentry{Listwise}
        \addplot[color=orange, mark=diamond*, dotted, mark options={scale=0.5, solid}]
            coordinates {
            (0.01, 0.037 )
            (0.1, 0.033 )
            (1, 0.032 )
            (10, 0.046 )
            (100, 0.012 )
            };
        % \addlegendentry{Listwise'}
        \end{axis}
        \end{tikzpicture}
        \caption{T-Learner}
    \end{subfigure}
    
    \vspace{5pt}
    
    \begin{subfigure}[b]{0.33\linewidth}
        \centering
        % \resizebox{\linewidth}{!}{%
        \begin{tikzpicture}
        \begin{axis}[
          axis lines=left,
          grid=both,
          grid style={line width=.1pt, draw=gray!10},
          major grid style={line width=0.5pt,draw=gray!30},
          xmin=0.01,
          xmax=100,
          xmode=log,
          ymin=0.0,
          ymax=0.075,
          xlabel=$\sigma$,
          ylabel=AUQC,
          x label style={at={(1, 0.10)}},
          ylabel style={rotate=-90},
          ylabel near ticks, 
          xtick={0.01, 0.1, 1, 10, 100},
          ytick={0.0, 0.02, ..., 0.4},
          hide obscured x ticks=false,
          hide obscured y ticks=false,
          width=\linewidth,
          height=\linewidth,
          legend style={
              draw=none, fill=gray!5, inner sep=2pt,
          },
          legend cell align={left},
          every axis plot/.append style={thick},
          y tick label style={
            /pgf/number format/.cd,
            fixed,
            fixed zerofill,
            precision=2,
            /tikz/.cd
          },
        ]
        % \footnotesize
        \addplot[color=green, mark=none, ]
            coordinates {
            (0.01, 0.054 )
            (100, 0.054 )
            };
        % \addlegendentry{Pointwise}
        \addplot[color=blue, mark=*, mark options={scale=0.5, solid}]
            coordinates {
            (0.01, 0.052 )
            (0.1, 0.053 )
            (1, 0.050 )
            (10, 0.050 )
            (100, 0.037 )
            };
        % \addlegendentry{Pairwise}
        \addplot[color=blue, mark=*, dotted, mark options={scale=0.5, solid}]
            coordinates {
            (0.01, 0.047 )
            (0.1, 0.050 )
            (1, 0.050 )
            (10, 0.045 )
            (100, 0.040 )
            };
        % \addlegendentry{Pairwise'}
        \addplot[color=orange, mark=diamond*, mark options={scale=0.5, solid}]
            coordinates {
            (0.01, 0.049 )
            (0.1, 0.051 )
            (1, 0.054 )
            (10, 0.049 )
            (100, 0.047 )
            };
        % \addlegendentry{Listwise}
        \addplot[color=orange, mark=diamond*, dotted, mark options={scale=0.5, solid}]
            coordinates {
            (0.01, 0.047 )
            (0.1, 0.046 )
            (1, 0.047 )
            (10, 0.043 )
            (100, 0.026 )
            };
        % \addlegendentry{Listwise'}
        \end{axis}
        \end{tikzpicture}
        % }
        \caption{X-Learner}
    \end{subfigure}
    \hfill
    \begin{subfigure}[b]{0.33\textwidth}
        \centering
        \begin{tikzpicture}
        \begin{axis}[
          axis lines=left,
          grid=both,
          grid style={line width=.1pt, draw=gray!10},
          major grid style={line width=0.5pt,draw=gray!30},
          xmin=0.01,
          xmax=100,
          xmode=log,
          ymin=0.0,
          ymax=0.075,
          xlabel=$\sigma$,
          ylabel=AUQC,
          x label style={at={(1, 0.10)}},
          ylabel style={rotate=-90},
          ylabel near ticks, 
          xtick={0.01, 0.1, 1, 10, 100},
          ytick={0.0, 0.02, ..., 0.4},
          hide obscured x ticks=false,
          hide obscured y ticks=false,
          width=\linewidth,
          height=\linewidth,
          legend style={
              draw=none, fill=gray!5, inner sep=2pt,
          },
          legend cell align={left},
          every axis plot/.append style={thick},
          y tick label style={
            /pgf/number format/.cd,
            fixed,
            fixed zerofill,
            precision=2,
            /tikz/.cd
          },
        ]
        % \footnotesize
        \addplot[color=green, mark=none, ]
            coordinates {
            (0.01, 0.056 )
            (100, 0.056 )
            };
        % \addlegendentry{Pointwise}
        \addplot[color=blue, mark=*, mark options={scale=0.5, solid}]
            coordinates {
            (0.01, 0.048 )
            (0.1, 0.050 )
            (1, 0.048 )
            (10, 0.048 )
            (100, 0.029 )
            };
        % \addlegendentry{Pairwise}
        \addplot[color=blue, mark=*, dotted, mark options={scale=0.5, solid}]
            coordinates {
            (0.01, 0.043 )
            (0.1, 0.043 )
            (1, 0.043 )
            (10, 0.047 )
            (100, 0.033 )
            };
        % \addlegendentry{Pairwise'}
        \addplot[color=orange, mark=diamond*, mark options={scale=0.5, solid}]
            coordinates {
            (0.01, 0.041 )
            (0.1, 0.047 )
            (1, 0.050 )
            (10, 0.045 )
            (100, 0.044 )
            };
        % \addlegendentry{Listwise}
        \addplot[color=orange, mark=diamond*, dotted, mark options={scale=0.5, solid}]
            coordinates {
            (0.01, 0.045 )
            (0.1, 0.041 )
            (1, 0.044 )
            (10, 0.046 )
            (100, 0.024 )
            };
        % \addlegendentry{Listwise'}
        \end{axis}
        \end{tikzpicture}
        \caption{DR-Learner}
    \end{subfigure}
    \hfill
    \begin{subfigure}[b]{0.33\textwidth}
    \centering
        \begin{tikzpicture}
        \begin{axis}[
          axis lines=left,
          grid=both,
          grid style={line width=.1pt, draw=gray!10},
          major grid style={line width=0.5pt,draw=gray!30},
          xmin=0.01,
          xmax=100,
          xmode=log,
          ymin=0.0,
          ymax=0.075,
          xlabel=$\sigma$,
          ylabel=AUQC,
          x label style={at={(1, 0.10)}},
          ylabel style={rotate=-90},
          ylabel near ticks, 
          xtick={0.01, 0.1, 1, 10, 100},
          ytick={0.0, 0.02, ..., 0.4},
          hide obscured x ticks=false,
          hide obscured y ticks=false,
          width=\linewidth,
          height=\linewidth,
          legend style={
              draw=none, fill=gray!5, inner sep=2pt,
          },
          legend cell align={left},
          every axis plot/.append style={thick},
          y tick label style={
            /pgf/number format/.cd,
            fixed,
            fixed zerofill,
            precision=2,
            /tikz/.cd
          },
        ]
        % \footnotesize
        \addplot[color=green, mark=none, ]
            coordinates {
            (0.01, 0.057 )
            (100, 0.057 )
            };
        % \addlegendentry{Pointwise}
        \addplot[color=blue, mark=*, mark options={scale=0.5, solid}]
            coordinates {
            (0.01, 0.051 )
            (0.1, 0.051 )
            (1, 0.047 )
            (10, 0.047 )
            (100, 0.038 )
            };
        % \addlegendentry{Pairwise}
        \addplot[color=blue, mark=*, dotted, mark options={scale=0.5, solid}]
            coordinates {
            (0.01, 0.038 )
            (0.1, 0.038 )
            (1, 0.038 )
            (10, 0.041 )
            (100, 0.045 )
            };
        % \addlegendentry{Pairwise'}
        \addplot[color=orange, mark=diamond*, mark options={scale=0.5, solid}]
            coordinates {
            (0.01, 0.049 )
            (0.1, 0.051 )
            (1, 0.052 )
            (10, 0.045 )
            (100, 0.037 )
            };
        % \addlegendentry{Listwise}
        \addplot[color=orange, mark=diamond*, dotted, mark options={scale=0.5, solid}]
            coordinates {
            (0.01, 0.039 )
            (0.1, 0.041 )
            (1, 0.037 )
            (10, 0.042 )
            (100, 0.033 )
            };
        % \addlegendentry{Listwise'}
        \end{axis}
        \end{tikzpicture}
        \caption{R-Learner}
    \end{subfigure}
\caption{\textit{What Is the Effect of the Sigmoid Parameter $\sigma$?} We show performance in terms of AUQC (higher is better) for the different metalearners on the \texttt{Synthetic} data set. We fix the number of sampling iterations $k = 1$ and train with default hyperparameters.}
\label{fig:experiment_synthetic_sigmoid}
\end{figure*}

In \cref{fig:experiment_synthetic_sigmoid}, we vary the sigmoid parametere $\sigma$, controlling the steepness of the comparison of the instance scores in the construction of the pairwise score (see \cref{eq:pairwise_score}). Generally, we obtain good performance for smaller values ($\sigma \leq 1$). When using normalization, our method seems more sensitive to this hyperparameter compared to training without score normalization. Although there may be some benefit of tuning this hyperparameter, we observe that fixing the sigmoid parameter at $\sigma = 1$ seems like a good choice overall--this was the setting used to generate the experimental results in the main body.

% SIGMOID

\end{document}